\documentclass[11pt]{article}
\usepackage[margin=1in]{geometry}
\usepackage{times}
\usepackage{natbib}

\usepackage{amsmath,amsfonts,bm}

\def\eqref#1{equation~\ref{#1}}

\def\1{\bm{1}}

\DeclareMathAlphabet{\mathsfit}{\encodingdefault}{\sfdefault}{m}{sl}
\SetMathAlphabet{\mathsfit}{bold}{\encodingdefault}{\sfdefault}{bx}{n}

\usepackage{amsmath, amssymb, amsfonts}
\usepackage{graphicx}
\usepackage{subcaption}
\usepackage{booktabs}
\usepackage{algorithm}
\usepackage{algpseudocode}
\usepackage{url}
\usepackage[colorlinks=true, allcolors=blue]{hyperref}

\title{\textbf{MetaLearnNCA: Few-Shot Offline Meta-Learning via\\Interacting Neural Cellular Automata}}

\author{
  Etienne Guichard \& Stefano Nichele \\
  Department of Computer Science and Communication \\
  \texttt{\{etienne.guichard,stefano.nichele\}@hiof.no}
}
\date{}

\begin{document}

\maketitle

\begin{abstract}
Few-shot meta-learning traditionally formulates task adaptation either as analytical gradient descent through unrolled computational graphs or as metric-based distance comparisons over flattened 1D feature vectors, which either incur costly test-time backpropagation or discard native 2D spatial geometry. In this work, we propose \textsc{MetaLearnNCA}, a decentralized framework that achieves few-shot adaptation through the dynamical interaction of coupled Neural Cellular Automata (NCAs) without computing analytical gradients during inference. \textsc{MetaLearnNCA} decomposes task adaptation into an \emph{Active-NCA}, which executes task inference conditioned on a continuous 2D spatial memory grid termed the \emph{spatial program}, and a learned \emph{Meta-NCA}, which acts as a decentralized cellular optimizer by diffusing spatial error residuals across local neighborhoods to dynamically update this program. \textsc{MetaLearnNCA} is competitive against canonical meta-learners in-distribution ($96.12\%$ on Omniglot) with  Out-Of-Distribution transfer gains on MNIST, KMNIST, and Fashion-MNIST transfer across 10 independent testing seeds across 1-, 5-, and 10-shot regimes (e.g., surpassing Prototypical Networks by $+10.54\%$ on 10-shot MNIST and a $+3.87\%$ gain on 10-shot Fashion-MNIST  over FOMAML). Our results establish that robust, gradient-free learning-to-learn can emerge from decentralized cellular dynamics on non-von Neumann substrates.
\end{abstract}

\section{Introduction}
\label{sec:introduction}

Few-shot meta-learning seeks to equip computational systems with the ability to acquire novel concepts rapidly from minimal supervision~\citep{hospedales2021meta, vinyals2016matching}. In biological organisms, this learning-to-learn capacity emerges not from monolithic, centralized gradient updates across billions of parameters, but through coordinated, decentralized cellular processes operating under strict physical and energetic constraints~\citep{levin2019computational}. Recently, \emph{Neural Cellular Automata} (NCAs)~\citep{mordvintsev2020growing} have emerged as a compelling paradigm of decentralized computing. Characterized by uniform local transition rules, translational invariance, and extreme parameter efficiency, NCAs have demonstrated remarkable prowess across morphogenesis~\citep{mordvintsev2020growing}, robust functional regeneration, texture synthesis ~\citep{niklasson2021self}, and spatial reasoning tasks—often utilizing orders of magnitude fewer parameters than conventional deep architectures.
\\

Crucially, recent investigations reveal that the representational capacity of NCAs extends beyond static pattern formation. Studies have demonstrated that NCAs organize their latent, non-observable channels into structured functional representations ~\citep{guichard2026neuralcellularautomatalearn, montero2026learningdevelopmentalscaffoldingsguide, randazzo2020self}. For instance, NCAs can learn reusable functional primitives in hidden state channels that are subsequently recombined to execute hierarchical, multi-stage behaviors~\citep{guichard2025engram, montero2026learningdevelopmentalscaffoldingsguide}. Furthermore, partitioning NCA states into private, immutable channels enables universal cellular computation, where distinct sub-channels act as isolated memory registers or programmable routing mechanisms~\citep{universalnca2025r}. In discriminative settings, NCAs have been shown to encode orthogonal, task-relevant representations within their latent channels; remarkably, features distilled from one NCA can bootstrap another to achieve competitive one-shot recognition~\citep{guichard2026neuralcellularautomatalearn}. This body of evidence points toward an unexplored frontier: \emph{the localized, recurrent hidden channels of NCAs provide a natural substrate for decentralized, gradient-free meta-learning.}
\\

In contrast, current deep meta-learning paradigms suffer from fundamental trade-offs: \emph{gradient-based} approaches~\citep{finn2017model, nichol2018reptile} require unrolling computational graphs and executing costly backpropagation during test-time inference, while \emph{black-box and in-context} methods~\citep{santoro2016meta, brown2020language} avoid test-time gradients only at the expense of massive parameter footprints ($10^8$--$10^{11}$ parameters), quadratic sequence complexity, and an absence of native 2D spatial inductive biases.
\\

To bridge this divide, we propose \textbf{MetaLearnNCA}, a decentralized framework that achieves few-shot offline adaptation without computing gradients at test time. MetaLearnNCA decomposes task adaptation into the dynamical interaction of two localized cellular systems:
\begin{itemize}
    \item An \textbf{Active-NCA}, which executes the primary task (e.g., visual character classification) conditioned on a localized, spatial memory grid termed the \emph{spatial program} 
    \item A \textbf{Meta-NCA}, which functions as a decentralized, learned cellular optimizer. Rather than calculating explicit gradients via chain-rule backpropagation, the Meta-NCA receives spatial residual error maps computed on the support set. Over a sequence of localized homogeneous transition functions, the Meta-NCA processes these residual errors and writes updated steering signals directly into the spatial program.
\end{itemize}

During adaptation, the Active-NCA and Meta-NCA communicate iteratively in a closed feedback loop: task inference produces residual errors, and cellular dynamics refines the spatial program. Once adaptation on the support set concludes, the resulting spatial program is frozen and broadcast across query instances. The query predictions are then computed purely through forward cellular unrolling—demanding \emph{zero gradient computation at adaptation or inference time}.
\\

\paragraph{Contributions.} (1) We introduce \textbf{\textsc{MetaLearnNCA}}, an offline meta-learning framework realizing few-shot adaptation via coupled NCAs that eliminates test-time backpropagation while maintaining a compact $<120\text{k}$ parameter footprint; (2) we formulate a \textbf{spatially distributed conditioning mechanism} where a learned Meta-NCA translates pixel-wise residual errors into persistent pre-activation program channels; and (3) we demonstrate \textbf{good empirical performance},with competetive performance on in-distribution ($96.12\%$ on Omniglot) with Out-Of-Distribution transfer gains on MNIST, KMNIST, and Fashion-MNIST.
    
\section{Related Work}
\label{sec:related_work}
Our work operates at the intersection of few-shot meta-learning, self-organizing Neural Cellular Automata, contrastive metric learning, and decentralized, biologically inspired credit assignment.

\subsection{Few-Shot Meta-Learning and Test-Time Adaptation}
Existing approaches in Few-shot Meta-Learning broadly divide into three paradigms: Gradient-Based Meta-Learning, Black-Box and Memory-Augmented Meta-Learning, and Metric-Based Learning.
\\

Optimization-centric algorithms, such as FOMAML \citep{finn2017model} and Reptile \citep{nichol2018reptile}, Meta-SGD \citep{li2017meta}, and implicit formulations \citep{rajeswaran2019implicit}, formalize meta-learning as optimizing a shared parameter initialization that adapts via inner-loop gradient descent on support examples. Although expressive, these methods suffer from computational issues: calculating second-order meta-gradients is memory-prohibitive, and test-time adaptation mandates analytical backpropagation through unrolled graphs.
\\

Memory-augmented architectures, such as MANN \citep{santoro2016meta} and SNAIL \citep{mishra2018simple}, replace gradient descent with recurrent dynamics, accumulating task representations directly in episodic memory. Specifically, MANN utilizes an external memory matrix with content-based addressing to dynamically store and recall class bindings, while SNAIL combines temporal causal convolutions with attention mechanisms to pinpoint past exemplars across an episode. However, both approaches flatten inputs into 1D feature vectors—discarding native 2D spatial geometry—and rely on centralized memory controllers or attention operations that scale poorly with episode length and introduce artificial order dependence into unordered support sets. In modern regimes, large language and vision models exhibit emergent in-context learning \citep{brown2020language, garg2022can, von2023transformers}. However, in-context inference typically requires massive parameter footprints ($10^8$–$10^{11}$), exhibits quadratic attention complexity, and lacks spatial locality priors.
\\

Methods such as Matching Networks \citep{vinyals2016matching}, Prototypical Networks \citep{snell2017prototypical}, and Relation Networks \citep{sung2018learning} map inputs into metric feature spaces where classification reduces to nearest-neighbor or prototype distance comparisons.

\subsection{Neural Cellular Automata: From Morphogenesis to General Computation}
Neural Cellular Automata (NCAs) integrate classical cellular automata \citep{von1966theory, wolfram2002new} with end-to-end differentiable neural components. Initially introduced to model biological morphogenesis, self-repair, and homeostatic pattern maintenance \citep{mordvintsev2020growing, niklasson2021self}, NCAs demonstrate how uniform, strictly localized update rules can coordinate globally stable anatomical structures.
\\

Beyond generative morphology, recent work has explored discriminative and computational NCA formulations. \citet{randazzo2020self} demonstrated that NCAs can achieve global agreement for digit classification (Self-classifying MNIST), proving that local message passing can perform perceptual classification without centralized pooling. Concurrently, efforts toward general and universal cellular computation \citep{variengien2021towards, universalnca2025r} have shown that continuous NCA state spaces can emulate discrete logic, matrix arithmetic, and routing operations. 
\\

To bridge low-level morphogenesis with high-level cognitive behavior, recent architectures have introduced modular memory and developmental hierarchies \citep{guichard2025engram, universalnca2025r, hei}. These architectures partition the hidden channel into mutable and immutable states, with the development of said states being decoupled from each other. Building upon this, NCAs have applied developmental cellular dynamics to the Abstraction and Reasoning Corpus (ARC-AGI) \citep{guichard2025arc, xu2025nca}, demonstrating that localized state-refinement dynamics are well-suited for abstract grid transformations. \textsc{MetaLearnNCA} advances this trajectory by formulating a dual-NCA interaction loop where an Active-NCA is modulated by a learned Meta-NCA that translates task-level residual signals into spatial program fields.
\\

\subsection{Decentralized and Biologically Plausible Credit Assignment}
Standard deep learning relies on global backpropagation, which exhibits well-documented biological implausibilities, including the weight-transport problem, non-local gradient flow, and global backward phases \citep{crick1989recent, lillicrap2020backpropagation}. Biologically plausible alternatives—such as Feedback Alignment \citep{lillicrap2016random}, Direct Feedback Alignment (DFA) \citep{nokland2016direct}, Predictive Coding \citep{ whittington2019theories}, and Equilibrium Propagation \citep{scellier2017equilibrium}—substitute exact backward transpose operations with local error projections or energy-relaxation phases.
\\

\section{Model Architecture}
\label{sec:model_architecture}
The \textbf{MetaLearnNCA} architecture performs few-shot spatial reasoning tasks through the dynamical coupling of two NCAs: an \textbf{Active-NCA} ($\mathbf{f}_\theta$), which executes task inference conditioned on a spatial program, and a \textbf{Meta-NCA} ($\mathbf{g}_\phi$), which iteratively infers and updates this spatial program by reading residual error signals. Both networks build upon the standard Neural Cellular Automata (NCA) formulation~\citep{mordvintsev2020growing} and conditioned routing mechanisms~\citep{guichard2025engram, universalnca2025r}.
\\

\subsection{Base NCA Formulation}
Both models share a common local transition backbone ($\text{BaseNCA}$). Let $\mathbf{s}^{(t)} \in \mathbb{R}^{B \times C_{\text{in}} \times H \times W}$ denote the cellular grid state at discrete time-step $t$. The local state transition comprises a fixed spatial perception operation followed by a point-wise Multi-Layer Perceptron (MLP):
\\

\paragraph{Perception Phase:} 
Each channel of the state grid is padded with circular boundary conditions and convolved with four fixed $3 \times 3$ differential filters $\mathbf{K} = \{K_{\text{ident}}, K_{\text{sobel}, x}, K_{\text{sobel}, y}, K_{\text{lap}}\}$, computing identity, directional Sobel gradients, and discrete Laplacian operators (exact filter matrices are given in Appendix~\ref{app:kernels}). Applying $\mathbf{K}$ across all $C_{\text{in}}$ channels yields a perceived feature representation $\mathbf{P}(\mathbf{s}^{(t)}) \in \mathbb{R}^{B \times 4C_{\text{in}} \times H \times W}$.
\\ 

\paragraph{Update Phase:}
The perceived vector per cell is mapped to a state differential $ds$ through a two-layer $1 \times 1$ convolutional network:
\begin{equation}
    ds = \mathbf{W}_2 \cdot \text{ReLU}\left(\mathbf{W}_1 \mathbf{P}(\mathbf{s}^{(t)}) + \mathbf{b}_1\right),
\end{equation}
where $\mathbf{W}_1 \in \mathbb{R}^{d_{\text{hidden}} \times 4C_{\text{in}}}$, $\mathbf{W}_2 \in \mathbb{R}^{C_{\text{out}} \times d_{\text{hidden}}}$ (with $d_{\text{hidden}} =256$), and $\mathbf{W}_2$ is initialized to zero without bias ($\mathbf{W}_2 = \mathbf{0}$) to ensure stable identity transitions at initialization. 
\\

To simulate asynchronous biological cellular behavior, cell updates are modulated by a stochastic dropout mask:
\begin{equation}
    \mathbf{s}_{\text{dyn}}^{(t+1)} = \mathbf{s}_{\text{dyn}}^{(t)} + ds \odot \mathbf{M}_{\text{stoch}}, \quad \text{where} \quad \mathbf{M}_{\text{stoch}} \sim \text{Bernoulli}(0.5) \in \{0, 1\}^{B \times 1 \times H \times W}.
\end{equation}

\subsection{Active NCA}
The Active-NCA $\mathbf{f}_\theta$ maps visual inputs to task solutions conditioned on a persistent spatial program $\mathbf{S}$. Its total state is partitioned into three functional sets ($C_{\text{in}} = 1 + C_s + C_d = 57$ channels): (i) \textbf{Observation Channel} ($C_{\text{img}} = 1$), the normalized input image $\mathbf{X} \in [-1, 1]^{B \times 1 \times H \times W}$; (ii) \textbf{Static Program Channels} ($C_s = 24$), the read-only spatial memory $\mathbf{S} \in \mathbb{R}^{B \times 24 \times H \times W}$ written by the Meta-NCA; and (iii) \textbf{Dynamic Reasoning Channels} ($C_d = 32$), the read-write state $\mathbf{D}_{\text{active}}$, where channels $0 \dots 9$ represent class logits maps $\hat{\mathbf{Y}}$ and channels $10 \dots 31$ serve as internal recurrent reasoning memory.
\\

Concatenating the static program grid $\mathbf{S}$ with the dynamic state prior to perception allows the program channels to serve as spatially distributed pre-activation biases on upstream convolutional neurons. Because each cell computes local state differentials based on its immediate neighborhood, this continuous spatial bias directly routes the flow of information across the grid during inference without requiring centralized routing circuitry (see Appendix~\ref{app:single_cell} and Figure~\ref{fig:cell_arch} for a detailed schematic of this conditioning mechanism).
\\

\subsection{Meta-NCA}
The Meta-NCA $\mathbf{g}_\phi$ acts as a decentralized optimizer. Rather than utilizing explicit gradient backpropagation during few-shot adaptation, it receives a spatial residual tensor $\mathbf{E}_{\text{pixel}} \in \mathbb{R}^{B \times C_{\text{loss}} \times H \times W}$ defined by:
\begin{equation}
    \mathbf{E}_{\text{pixel}} = (\mathbf{Y}_{\text{spatial}} - \text{Softmax}(\hat{\mathbf{Y}})) \odot \mathbf{M}_{\text{alive}},
\end{equation}
where $\mathbf{Y}_{\text{spatial}} \in \{0, 1\}^{B \times 10 \times H \times W}$ is the one-hot target tensor broadcast spatially across all pixels, and $\mathbf{M}_{\text{alive}} = \mathbb{I}(\mathbf{X} > -0.8)$ is a binary foreground mask that isolates character strokes and prevents background pixels from generating penalizing error signals. The masking is also used during training when computing class logits, as in ~\citet{randazzo2020self}.
\\

The Meta-NCA state combines $\mathbf{E}_{\text{pixel}}$ with its internal dynamic memory $\mathbf{D}_{\text{meta}} \in \mathbb{R}^{B \times 32 \times H \times W}$ ($C_{\text{in}} = 10 + 32 = 42$ channels). Over $N$ internal recurrence steps, the Meta-NCA diffuses local error feedback across neighboring cells to update $\mathbf{D}_{\text{meta}}$. The updated task program is extracted directly from the first $C_s = 24$ dynamic channels.
\\

\subsection{Ensemble Interaction Loop}
The full MetaLearnNCA framework coordinates the two models within a nested interaction loop (Figure~\ref{fig:MetaLearnNCA}; formal algorithmic outline provided in Appendix~\ref{app:algo}, Algorithm~\ref{alg:meta_nca_training}):
\begin{enumerate}
    \item \textbf{Initialization:} A shared spatial program grid is initialized randomly: $\mathbf{S}_{\text{task}} \sim \mathcal{U}(0, 1)^{1 \times 24 \times H \times W}$, and the meta dynamic state is cleared ($\mathbf{D}_{\text{meta}} \leftarrow \mathbf{0}$).
    \item Support Adaptation ($K$ iterations): For each interaction step $k \in \{1, \dots, K\}$:
    \begin{enumerate}
        \item Broadcast $\mathbf{S}_{\text{task}}$ across the support batch: $\mathbf{S} \leftarrow \text{expand}(\mathbf{S}_{\text{task}}, B_{\text{supp}})$.
        \item Execute Active-NCA for $2N$ steps: $\mathbf{D}_{\text{supp}} \leftarrow \mathbf{f}_\theta(\mathbf{X}_{\text{supp}}, \mathbf{S}, \text{steps} = 2N)$.
        \item Compute the residual $\mathbf{E}_{\text{pixel}}$ over the support set.
        \item Execute Meta-NCA for $N$ steps: $\mathbf{D}_{\text{meta}} \leftarrow \mathbf{g}_\phi( \mathbf{E}_{\text{pixel}}, \mathbf{D}_{\text{meta}}, \text{steps} = N)$.
        \item Aggregate support programs via batch mean-pooling: $\mathbf{S}_{\text{task}} \leftarrow \frac{1}{B_{\text{supp}}} \sum_{i=1}^{B_{\text{supp}}} \mathbf{D}_{\text{meta}, i}^{[0:C_s, :, :]}$.
    \end{enumerate}
    \item \textbf{Query Inference:} The adapted program $\mathbf{S}_{\text{task}}$ is frozen and broadcast to the query set: $\mathbf{S}_{\text{qry}} \leftarrow \text{expand}(\mathbf{S}_{\text{task}}, B_{\text{qry}})$. The Active-NCA then executes inference on $\mathbf{X}_{\text{qry}}$ for $2N$ steps to produce the final predictions $\mathbf{D}_{\text{qry}}$.
\end{enumerate}
\textbf{Hypothetical Mechanics of Distributed Memory Superposition.} Unlike metric meta-learners that store isolated 1D prototype vectors, \textsc{MetaLearnNCA} encodes the full multi-class task representation within a single, superposed spatial memory manifold $S_{\text{task}} \in \mathbb{R}^{C_s \times H \times W}$. In high-dimensional cellular lattices ($C_s \times H \times W = 18{,}816$), class-specific steering fields written by the Meta-NCA could occupy quasi-orthogonal subspaces. Batch mean-pooling in step (e) could acts as an additive holographic superposition. This was not empirically tested but could explain why batch mean pooling does not destructively interfere with class specific artifacts.

\begin{figure}[t]
    \centering
    \includegraphics[width=0.75\linewidth]{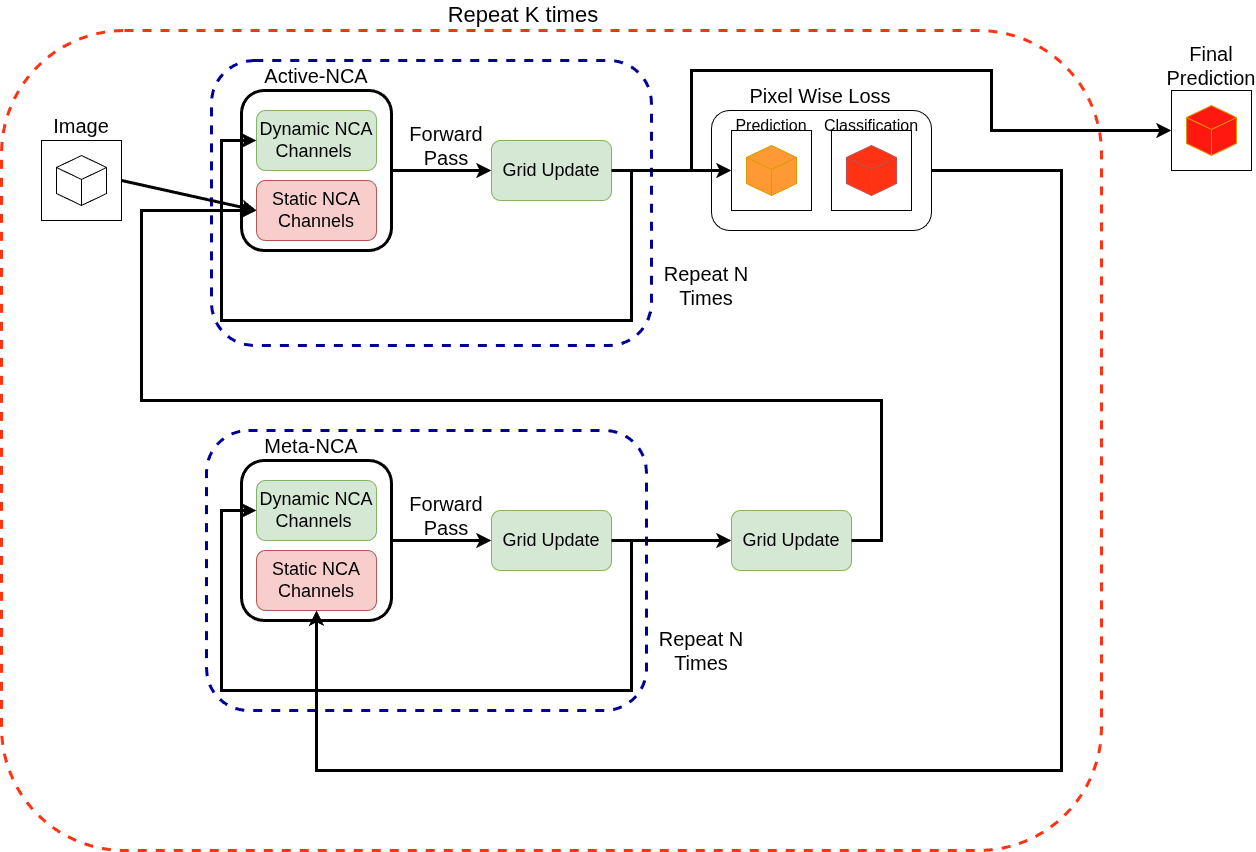}
    \caption{The \textbf{MetaLearnNCA} interaction loop: Active-NCA computes forward task predictions on the support set, while Meta-NCA translates the residual error maps into an updated spatial program $\mathbf{S}_{\text{task}}$.}
    \label{fig:MetaLearnNCA}
\end{figure}

\section{Training Methodology}
The parameters $\Theta = \{\theta, \phi\}$ of both the Active-NCA and Meta-NCA are optimized concurrently through the unrolled adaptation trajectory (Algorithm~\ref{alg:meta_nca_training}, Appendix~\ref{app:algo}), heavily inspired by FOMAML~\citep{finn2017model}.

\paragraph{Meta-Objective Formulation:} The query loss $\mathcal{L}_{\text{qry}}$ is computed as a stroke-masked spatial Cross-Entropy loss. For each query instance $i$, raw logit maps $\hat{\mathbf{Y}}_{\text{qry}, i} \in \mathbb{R}^{C_{\text{loss}} \times H \times W}$ are supervised by the scalar ground-truth label $y_{\text{qry}, i}$ broadcast spatially across all grid coordinates $(u, v)$ belonging to the active stroke mask $\mathbf{M}_{\text{qry}, i}$:
\begin{equation}
\mathcal{L}_{\text{qry}} = \frac{\sum_{i=1}^{B_{\text{qry}}} \sum_{u=1}^H \sum_{v=1}^W -\log\left(\frac{\exp(\hat{\mathbf{Y}}_{\text{qry}, i}(y_{\text{qry}, i}, u, v))}{\sum_{c=0}^{C_{\text{loss}}-1}\exp(\hat{\mathbf{Y}}_{\text{qry}, i}(c, u, v))}\right) \cdot \mathbf{M}_{\text{qry}, i}(u, v)}{\sum_{i=1}^{B_{\text{qry}}} \sum_{u=1}^H \sum_{v=1}^W \mathbf{M}_{\text{qry}, i}(u, v) + \epsilon}
\end{equation}
During discrete test-time evaluation, predictions $\hat{y}_i$ are extracted by globally average pooling logits across all active stroke pixels:
\begin{equation}
\label{eq:eval_pooling}
\hat{y}_i = \arg\max_{c \in \{0, \dots, C_{\text{loss}}-1\}} \frac{\sum_{u=1}^H \sum_{v=1}^W \hat{\mathbf{Y}}_{\text{qry}, i}(c, u, v) \cdot \mathbf{M}_{\text{qry}, i}(u, v)}{\sum_{u=1}^H \sum_{v=1}^W \mathbf{M}_{\text{qry}, i}(u, v) + \epsilon}.
\end{equation}
\\

\paragraph{Temporal Jittering for Robustness:}
To prevent the cellular dynamics from overfitting to exact unroll lengths, we apply random step-length jittering during meta-training ($N' \in [16, 19]$ and $K' \in [3, 5]$). The total effective minimal interaction steps ($K \times 3N = 3 \times 48 = 144$) substantially exceed the canvas diameter ($H=W=28$, grid diagonal $\approx 39.6$), ensuring sufficient time for information to propagate across NCAs (with kernels of size 3, the speed of light within an NCA is 1 pixel per step).
\\

\section{Experiments}
We evaluate MetaLearnNCA on few-shot character recognition using the Omniglot dataset ($964$ background classes) for training. All images are resized to $28 \times 28$, inverted such that background pixels equal $0.0$ and character strokes equal $1.0$, and normalized to $[-1, 1]$.

\paragraph{Episode Sampling:}
In each meta-training iteration, we randomly sample a 10-way, 1-support, 5-query task ($B_{\text{supp}} = 10$, $B_{\text{qry}} = 50$). True Omniglot class IDs are mapped dynamically to surrogate class indices $\{0, \dots, 9\}$. 

\paragraph{Optimization Details:}
The joint model is trained for $E = 5000$ meta-epochs with a meta-batch size of $B = 32$ tasks using AdamW ($lr = 10^{-3}$, weight decay $= 10^{-5}$). Gradients are clipped to a maximum $\ell_2$ norm of $1.0$.

\section{Results}
We evaluate \textsc{MetaLearnNCA} along three dimensions: (i) few-shot out-of-distribution (OOD) cross-dataset transfer, (ii) spatial fault tolerance and damage resilience, and (iii) qualitative visual inspection of the learned spatial program fields. We present further results and analysis in Appendices \ref{app:hcspa}, \ref{app:fp}, \ref{app: ssmc}, \ref{app:d2dt}, \ref{app:kakmls} \ref{sec:spatial_translation_robustness}.

\subsection{Out Of Distribution Offline Learning}
To evaluate whether the learned cellular optimizer (\textsc{Meta-NCA}) acquires a generalizable error-diffusion dynamic rather than memorizing in-distribution character templates, we evaluate models meta-trained solely on Omniglot across five completely unseen out-of-distribution visual domains: \textbf{MNIST}, \textbf{USPS}, \textbf{KMNIST}, \textbf{Fashion-MNIST}, and \textbf{MedMNIST Organ} (referred to as MedMNIST). Full testing algorithm can be found in Appendix \ref{sec:test_algo}.
\\

\begin{table}[t]
\caption{Few-shot adaptation performance (Accuracy \% $\pm$ 95\% CI) across 1-shot, 5-shot, and 10-shot regimes on in-domain Omniglot and cross-domain out-of-distribution transfer benchmarks. All models (except ours) employ the canonical Conv4 backbone on $28\times 28$ inputs.}
\label{tab:main_benchmark_comparison}
\vskip 0.1in
\centering
\footnotesize
\setlength{\tabcolsep}{2.6pt}
\begin{tabular}{lc cccc | c}
\toprule
 & & \multicolumn{3}{c}{\textbf{Gradient-Free Baselines}} & \textbf{Gradient-Based} & \textbf{Ours} \\
\textbf{Dataset} & \textbf{Shot} & \textbf{MatchingNet} & \textbf{RelationNet} & \textbf{ProtoNet} & \textbf{FOMAML} & \textbf{MetaLearnNCA} \\
\midrule
{\textbf{Omniglot (In-Dist)}} 
 & 1  & $63.47 \pm 7.79$ & $84.84 \pm 3.74$ & $\mathbf{92.26 \pm 3.15}$ & $87.84 \pm 3.67$ & $87.95 \pm 4.16$ \\
 & 5  & $75.47 \pm 4.54$ & $92.73 \pm 3.09$ & $96.40 \pm 3.09$ & $\mathbf{96.93 \pm 1.41}$ & $95.20 \pm 2.13$ \\
 & 10 & $77.40 \pm 6.37$ & $93.20 \pm 2.65$ & $97.40 \pm 1.27$ & $\mathbf{98.80 \pm 0.94}$ & $96.12 \pm 0.70$ \\
\midrule
{\textbf{MNIST}} 
 & 1  & $44.79 \pm 4.93$ & $51.17 \pm 2.66$ & $55.32 \pm 1.87$ & $48.16 \pm 2.98$ & $\mathbf{65.02 \pm 3.07}$ \\
 & 5  & $58.93 \pm 1.34$ & $63.74 \pm 1.39$ & $75.17 \pm 1.56$ & $70.80 \pm 1.72$ & $\mathbf{84.12 \pm 1.57}$ \\
 & 10 & $58.66 \pm 1.15$ & $67.30 \pm 0.66$ & $79.17 \pm 1.48$ & $72.37 \pm 2.01$ & $\mathbf{89.71 \pm 0.43}$ \\
\midrule
{\textbf{USPS}} 
 & 1  & $30.23 \pm 3.33$ & $24.14 \pm 3.46$ & $35.50 \pm 3.74$ & $42.40 \pm 4.62$ & $\mathbf{53.26 \pm 5.99}$ \\
 & 5  & $34.39 \pm 3.03$ & $35.41 \pm 2.69$ & $51.07 \pm 2.04$ & $\mathbf{64.91 \pm 2.13}$ & $57.10 \pm 4.26$ \\
 & 10 & $36.58 \pm 3.02$ & $38.81 \pm 1.19$ & $55.54 \pm 1.89$ & $\mathbf{69.75 \pm 2.16}$ & $65.02 \pm 3.10$ \\
\midrule
{\textbf{Fashion-MNIST}} 
 & 1  & $25.24 \pm 2.84$ & $20.82 \pm 2.59$ & $22.26 \pm 1.75$ & $31.39 \pm 3.21$ & $\mathbf{40.43 \pm 3.11}$ \\
 & 5  & $31.68 \pm 0.82$ & $25.43 \pm 0.75$ & $29.66 \pm 0.91$ & $47.41 \pm 1.88$ & $\mathbf{51.34 \pm 2.67}$ \\
 & 10 & $32.99 \pm 0.75$ & $25.91 \pm 1.58$ & $31.63 \pm 1.29$ & $50.53 \pm 1.82$ & $\mathbf{54.40 \pm 1.22}$ \\
\midrule
{\textbf{KMNIST}} 
 & 1  & $17.15 \pm 1.46$ & $20.04 \pm 1.96$ & $22.57 \pm 2.35$ & $22.76 \pm 2.01$ & $\mathbf{30.28 \pm 3.14}$ \\
 & 5  & $23.15 \pm 1.03$ & $26.06 \pm 1.60$ & $30.09 \pm 2.05$ & $33.66 \pm 1.66$ & $\mathbf{41.65 \pm 1.86}$ \\
 & 10 & $25.61 \pm 1.38$ & $29.40 \pm 1.13$ & $35.42 \pm 1.42$ & $36.23 \pm 1.18$ & $\mathbf{51.26 \pm 1.27}$ \\
\midrule
{\textbf{MedMNIST}} 
 & 1  & $17.47 \pm 3.00$ & $10.79 \pm 3.83$ & $20.86 \pm 2.66$ & $\mathbf{22.82 \pm 4.74}$ & $18.54 \pm 2.16$ \\
 & 5  & $26.61 \pm 3.19$ & $12.24 \pm 3.51$ & $24.85 \pm 2.34$ & $\mathbf{35.91 \pm 2.30}$ & $19.76 \pm 3.86$ \\
 & 10 & $27.88 \pm 2.80$ & $11.86 \pm 4.70$ & $25.99 \pm 2.14$ & $\mathbf{34.24 \pm 1.31}$ & $26.43 \pm 2.22$ \\
\midrule
\multicolumn{2}{l}{\textbf{Test-Time Gradients?}} & \textbf{No} & \textbf{No} & \textbf{No} & \textit{Yes (5 steps)} & \textbf{No} \\
\multicolumn{2}{l}{\textbf{Parameters}} & $\approx 115\text{k}$ & $\approx 225\text{k}$ & $\approx 115\text{k}$ & $\approx 115\text{k}$ & $\mathbf{\approx 118\text{k}}$ \\
\bottomrule
\end{tabular}
\end{table}

Table~\ref{tab:main_benchmark_comparison} evaluates \textsc{MetaLearnNCA} against canonical metric- and optimization-based baselines across 1-shot, 5-shot, and 10-shot adaptation regimes over 10 independent test splits (1 random checkpoint, 10 testing seeds). While FOMAML and ProtoNet maintain a marginal lead in-domain on Omniglot ($\approx 97\%$--$98\%$ at 10-shot). In contrast, \textsc{MetaLearnNCA} demonstrates somewhat better cross-domain transfer: it outperforms all baselines across \emph{all} shot counts on MNIST (reaching $\mathbf{89.71\%}$ at 10-shot, $+10.54\%$ over ProtoNet), cursive KMNIST ($\mathbf{51.26\%}$, $+15.03\%$ over FOMAML), and Fashion-MNIST ($\mathbf{54.40\%}$, $+3.87\%$ over FOMAML). Furthermore, in a 1-shot domain setting, \textsc{MetaLearnNCA} achieves the highest accuracy across 4 of 5 unseen target domains (including $\mathbf{53.26\%}$ on USPS, beating FOMAML by $+10.86\%$). FOMAML outperformed MetaLearnNCA across all shots on MedMNIST ($+4.28\%$, $+16.51\%$, $+7.81\%$) and on 1- and 5- shot on USPS ($+7.81\%$, $+4.72\%$). Crucially, \textsc{MetaLearnNCA} achieves this transfer purely through forward cellular dynamics.
\\

\begin{figure}[t]
    \centering
    \includegraphics[width=0.72\linewidth]{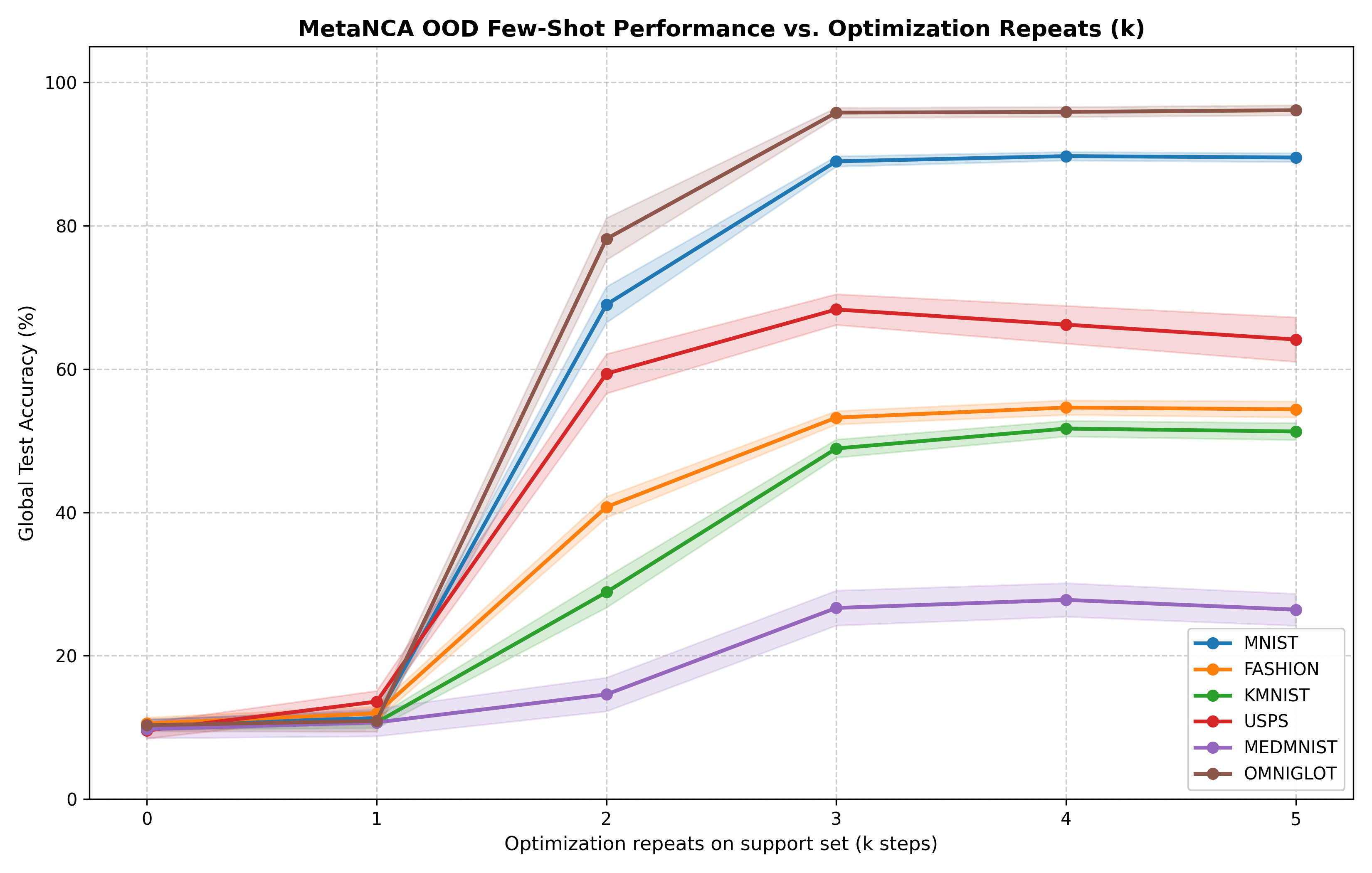}
    \caption{Few-shot offline 10-shot adaptation trajectories of \textsc{MetaLearnNCA} evaluated on out-of-distribution character datasets across offline cellular optimization repeats ($k \in \{0, \dots, 5\}$). Solid curves depict mean accuracy across 42 independent meta-training runs and 42 independent testing seeds, with shaded regions denoting 95\% confidence intervals.}
    \label{fig:results1}
\end{figure}

To measure end-to-end (training and testing) variability, Figure~\ref{fig:results1} plots test-time adaptation trajectories across offline cellular optimization repeats ($k \in \{0, \dots, 5\}$) in the 10-shot regime, aggregated across $\mathbf{42}$ independent meta-training runs and $\mathbf{42}$ independent testing seeds (One random test seed per random training seed). Crucially, the 95\% confidence interval bands demonstrate that \textsc{MetaLearnNCA} converges consistently regardless of the training or testing seed. Across all domains, the coupled cellular dynamics display an emergent two-phase behavior: remaining flat at chance level ($\approx 10\%$) at $k=0$ and $k=1$, followed by a sharp transition at $k=2$, and plateauing stably between $k=3$ and $k=4$ without diverging under extended recurrence (except for USPS).
\\

\subsection{Program Damage}
A defining characteristic of biological cellular ensembles is structural fault tolerance. We evaluate whether the continuous spatial representation $S_{\text{task}}$ retains this physical property under severe localized damage.
\\

Following support adaptation, we corrupt the spatial program grid $S_{\text{task}}$ by injecting circular patches of random noise of radius $r = 4\text{ pixels}$ ($A_{\text{damage}} \approx 50.3\text{ px}^2$, spanning $\approx 6.4\%$ of the total $28\times 28$ grid area per circle across all 24 channels). We evaluate test classification accuracy as a function of the number of damaged patches ($d \in \{0, \dots, 6\}$).
\\

\begin{figure}[t]
    \centering
    \includegraphics[width=0.72\linewidth]{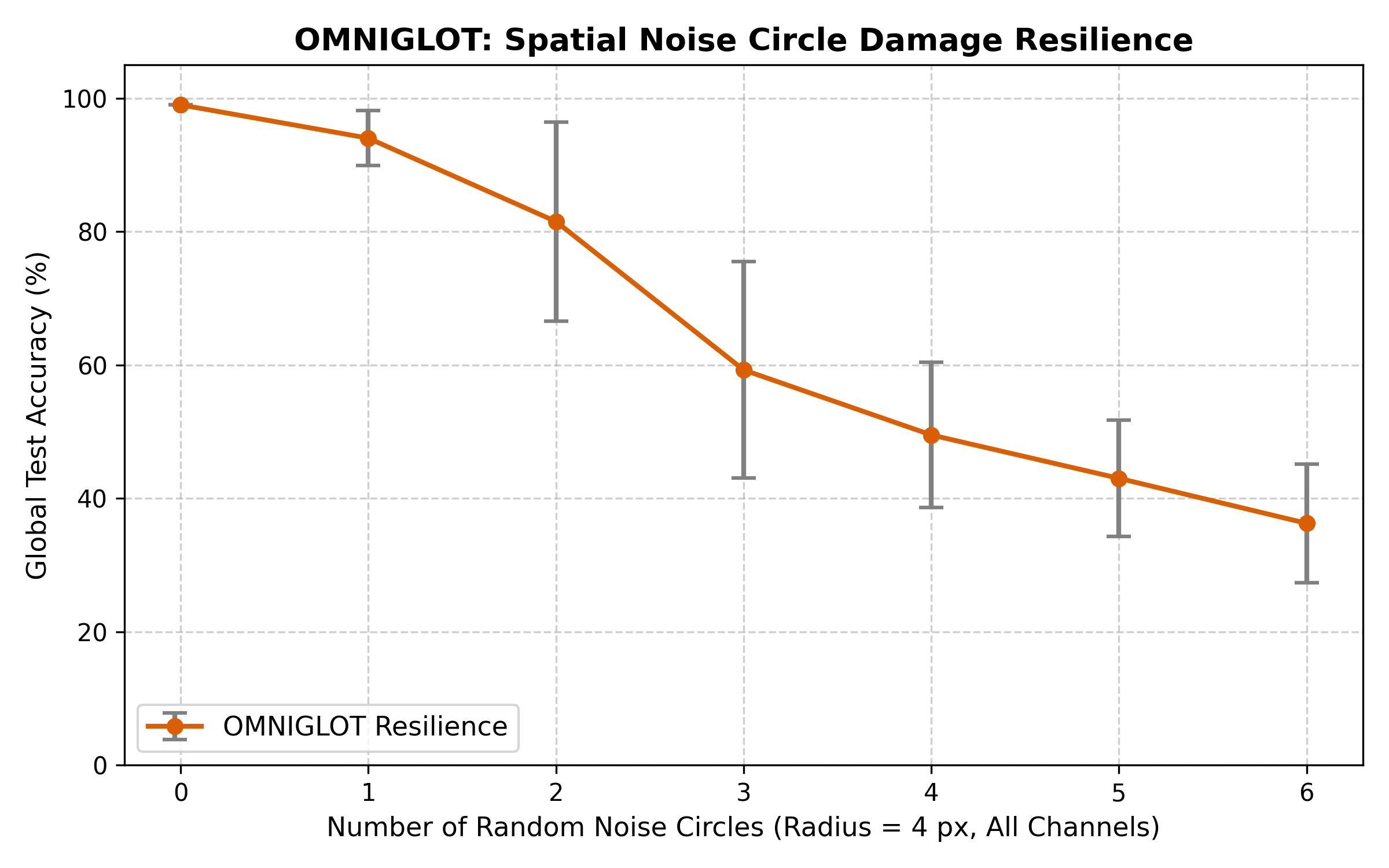}
    \caption{Resilience of MetaLearnNCA to physical spatial damage: test accuracy as a function of circular noise patches ($r = 4\text{ px}$) injected into the static program grid $S_{\text{task}}$.}
    \label{fig:prog_dam}
\end{figure}

As depicted in Figure~\ref{fig:prog_dam}, with a single localized circular puncture ($d=1$), \textsc{MetaLearnNCA} retains $\mathbf{93.4\%}$ classification accuracy (a marginal $\approx 4.6\%$ drop relative to the uncorrupted $98.0\%$ baseline). Even when subjected to two non-overlapping punctures ($d=2$, destroying over $12\%$ of the spatial memory footprint), performance remains resilient at an average $\mathbf{81.8\%}$. Variance between runs, however, increases greatly. Performance degrades smoothly as damage scales ($d \ge 3$), dropping toward chance only when larger fractions of the grid are physically destroyed.

\subsection{Hidden Channel and Program Field Visualizations}

We visualize a small section of the hidden channels and program on the MNIST dataset. We chose to visualize MNIST as the characters are easier to recognize.
\\

\begin{figure}[t]
    \centering
    \includegraphics[width=0.92\linewidth]{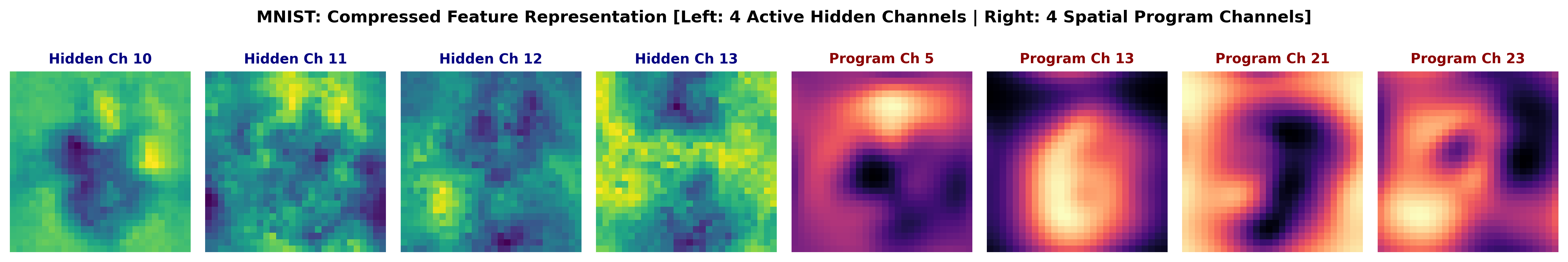}
    \caption{Representative spatial and program channels during MNIST inference. (see Appendix~\ref{app:atlases} for the complete 46-channel atlas).}
    \label{fig:representative_channels}
\end{figure}
Figure~\ref{fig:representative_channels} illustrates representative hidden reasoning channels from the Active-NCA and program channels from the Meta-NCA during inference on Mnist. The dynamic reasoning channels exhibit chaotic behavior, possibly representing a higher-order feature of the classification task. The program channels demonstrate functional specializations such as: Spot attention (channel 5), positive general shape matching (channel 13), negative shape matching (channel 21), and more complex compound features (channel 23). See Appendix~\ref{app:atlases} for the complete 46-channel atlas.
\section{Discussion}
\label{sec:discussion}

Beyond eliminating test time gradients, \textsc{MetaLearnNCA} shifts the representational paradigm of few-shot learning by encoding task knowledge as a continuous, spatially structured dynamical field rather than parametric weight perturbations or global 1D vector embeddings. Crucially, our framework embodies a decentralized realization of localized credit assignment that bypasses the non-local weight-transport requirements of backpropagation. Instead of back-propagating errors through analytical layer inversion, task adaptation modifies the spatial states used by the networks during forward computation, while the weights remain fixed. This demonstrates how effective task adaptation can emerge on non-Von Neumann substrates with explicit aggregation operations.
\\

From a neurobiological perspective, \textsc{MetaLearnNCA} can be viewed as a speculative spatially grounded abstraction of biological cortical dynamics. Each cell on the lattice functions analogously to an individual neuron or cortical microcolumn, with a local $3 \times 3$ perception that models the recurrent connectivity among local neurons. Within this microcircuit, dynamic reasoning channels emulate fast-timescale electrophysiological variables such as membrane potentials and intracellular signaling cascades, while the persistent spatial program channels ($S$) operate as slow neuromodulatory fields or synaptic trace maps that alter upstream neural excitability without modifying structural weights. In this view, the Meta-NCA acts as a feedback or subcortical neuromodulatory network that detects sensory prediction errors and broadcasts chemical or electrical signals throughout the tissue, thereby regulating the attractor dynamics of the primary network entirely through recurrent activations.
\\

Several limitations of this study highlight important avenues for future work. First, our evaluation relies on character and silhouette benchmarks where visual concepts exhibit a strong spatial center bias, which naturally aids coordinate-dependent steering; extending this formulation to unconstrained scenes with translation, scale variance, and background clutter remains to be demonstrated. Second, our core architectural hyperparameters—including channel partitions ($C_s=24, C_d=32$), MLP hidden widths ($d_{\text{hidden}}=256$), inner steps ($N=16$), and outer test repeats ($K=5$)—were selected heuristically rather than through systematic optimization sweeps. Third, while test-time adaptation is gradient-free and activation memory is independent of unroll length during inference, unrolling recurrent trajectories makes meta-training computationally demanding in wall-clock time and backpropagation memory. The mechanism for modulation (appending the program channels to the NCA) remains limited in functional expressivity. Scaling directly to higher-resolution RGB canvases is constrained by the cellular speed of light. Finally, the experiments do not isolate the benefits of the cellular updater, spatial memory, and foreground masking, or establish a runtime advantage. Future work will investigate Graph Neural Cellular Automata (GNCAs) over flexible topological graphs, developing methods for more expressive programs, and apply interacting NCAs to abstract program induction on the Abstraction and Reasoning Corpus (ARC-AGI).

\section{Conclusion}
\label{sec:conclusion}
We introduced \textbf{MetaLearnNCA}, an offline meta-learning framework that achieves few-shot adaptation through the dynamical interaction of coupled Neural Cellular Automata without test-time backpropagation. By formulating task representations as continuous 2D spatial memory fields and computing credit assignment via localized updates to a residual error map, \textsc{MetaLearnNCA} bridges the gap between gradient-free meta-learning and decentralized cellular self-organization. Our framework matches established baselines in-distribution while achieving gains on some OOD transfer tasks. \textsc{MetaLearnNCA} demonstrates that gradient-free learning-to-learn can emerge from decentralized cellular dynamics on non-von Neumann computing substrates.

\subsection*{AI use statement}
In this work, generative AI assistance was used for LaTeX syntax restructuring, template formatting, and phrasing refinement. All theoretical derivations, algorithmic designs, experimental implementations, and conclusions were conceived, verified, and validated by the author(s).

\subsubsection*{Reproducibility Statement}
 Complete PyTorch code, training scripts, and evaluation routines for both \textsc{MetaLearnNCA} and all baselines are open-sourced via an anonymous repository at: \url{https://github.com/etimush/MetaLearnNCA}.

\subsection*{Acknowledgment}
This work was supported by MishMash - Research Council of Norway, GrantId: 357438.

\bibliographystyle{plainnat}
\bibliography{iclr2027_conference}

\newpage
\appendix

\section{Extended Architectural and Algorithmic Details}
\label{app:arch}

\subsection{Explicit Perception Filter Matrices}
\label{app:kernels}
The four fixed $3 \times 3$ differential filters $\mathbf{K} = \{K_{\text{ident}}, K_{\text{sobel}, x}, K_{\text{sobel}, y}, K_{\text{lap}}\}$ used in the perception phase (Section 3.1) are defined as:
\begin{equation}
    K_{\text{ident}} = \begin{bmatrix} 0 & 0 & 0 \\ 0 & 1 & 0 \\ 0 & 0 & 0 \end{bmatrix}, \quad
    K_{\text{sobel}, x} = \begin{bmatrix} -1 & 0 & 1 \\ -2 & 0 & 2 \\ -1 & 0 & 1 \end{bmatrix}, \quad
    K_{\text{sobel}, y} = K_{\text{sobel}, x}^\top, \quad
    K_{\text{lap}} = \begin{bmatrix} 1 & 1 & 1 \\ 1 & -8 & 1 \\ 1 & 1 & 1 \end{bmatrix}.
\end{equation}

\subsection{Single-Cell Conditioning Schematic}
\label{app:single_cell}
\begin{figure}[h]
    \centering
    \includegraphics[width=0.65\linewidth]{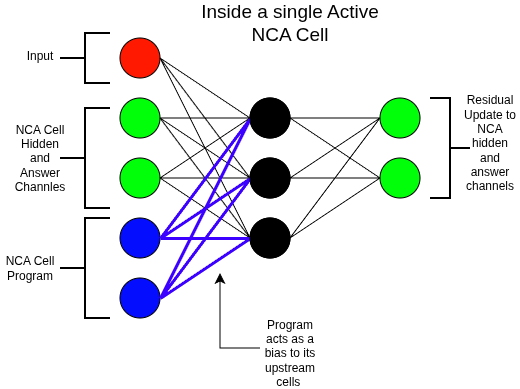}
    \caption{Theoretical overview of the conditioning mechanism: the static spatial program $\mathbf{S}$ acts as a localized pre-activation bias, dynamically modulating information flow through the Active-NCA transition operator.}
    \label{fig:cell_arch}
\end{figure}

\subsection{Meta-NCA Meta-Training Algorithm}
\label{app:algo}

\begin{algorithm}[H]
\caption{Meta-NCA Meta-Training Loop}
\label{alg:meta_nca_training}
\begin{algorithmic}[1]
    \Require Dataset $\mathcal{D}$, Meta-batch size $B$, Epochs $E$, Base Inner iterations $K=3$, Base inner steps $N=16$, Learning rate $\eta=10^{-3}$
    \State Initialize Active-NCA $\mathbf{f}_\theta$ and Meta-NCA $\mathbf{g}_\phi$
    \State Initialize AdamW optimizer for joint parameters $\Theta = \{\theta, \phi\}$

    \For{$\text{epoch} = 1 \text{ to } E$}
        \State $\nabla_\Theta \mathcal{L} \leftarrow 0$
        
        \For{$\text{task } b = 1 \text{ to } B$}
            \State Sample episode $(\mathbf{X}_{\text{supp}}, \mathbf{y}_{\text{supp}}), (\mathbf{X}_{\text{qry}}, \mathbf{y}_{\text{qry}}) \sim \mathcal{D}$
            \State Initialize program $\mathbf{S}_{\text{task}} \sim \mathcal{U}(0, 1)^{1 \times C_s \times H \times W}$ and meta-state $\mathbf{D}_{\text{meta}} \leftarrow \mathbf{0}$
            \State Sample randomized step horizons: $N' \sim \mathcal{U}\{N, N+3\}$, $K' \sim \mathcal{U}\{K, K+2\}$
            
            \Statex \quad\quad \textbf{// Inner Adaptation Loop (Support Set)}
            \For{$k = 1 \text{ to } K'$}
                \State $\mathbf{S} \leftarrow \text{expand}(\mathbf{S}_{\text{task}}, B_{\text{supp}})$
                \State $\mathbf{D}_{\text{supp}} \leftarrow \mathbf{f}_\theta(\mathbf{X}_{\text{supp}}, \mathbf{S}, \text{steps}=2N')$
                \State $\mathbf{E}_{\text{pixel}} \leftarrow \left(\mathbf{Y}_{\text{supp}} - \text{Softmax}(\mathbf{D}_{\text{supp}}^{[0:C_{\text{loss}}]})\right) \odot \mathbb{I}(\mathbf{X}_{\text{supp}} > -0.8)$
                \State $\mathbf{D}_{\text{meta}} \leftarrow \mathbf{g}_\phi( \mathbf{E}_{\text{pixel}}, \mathbf{D}_{\text{meta}}, \text{steps}=N')$
                \State $\mathbf{S}_{\text{task}} \leftarrow \frac{1}{B_{\text{supp}}} \sum_{i=1}^{B_{\text{supp}}} \mathbf{D}_{\text{meta}, i}^{[0:C_s]}$
            \EndFor
            
            \Statex \quad\quad \textbf{// Query Evaluation \& Meta-Loss}
            \State $\mathbf{S}_{\text{qry}} \leftarrow \text{expand}(\mathbf{S}_{\text{task}}, B_{\text{qry}})$
            \State $\mathbf{D}_{\text{qry}} \leftarrow \mathbf{f}_\theta(\mathbf{X}_{\text{qry}}, \mathbf{S}_{\text{qry}}, \text{steps}=2N)$
            \State $\hat{\mathbf{Y}}_{\text{qry}} \leftarrow \mathbf{D}_{\text{qry}}^{[0:C_{\text{loss}}]}$, \quad $\mathbf{M}_{\text{qry}} \leftarrow \mathbb{I}(\mathbf{X}_{\text{qry}} > -0.8)$
            \State Compute masked spatial Cross-Entropy:
            \Statex \qquad\qquad $\mathcal{L}_{\text{qry}} \leftarrow \frac{\sum_{i=1}^{B_{\text{qry}}} \sum_{u=1}^H \sum_{v=1}^W \ell_{\text{CE}}\left(\hat{\mathbf{Y}}_{\text{qry}, i}(\cdot, u, v),\, y_{\text{qry}, i}\right) \cdot \mathbf{M}_{\text{qry}, i}(u, v)}{\sum_{i=1}^{B_{\text{qry}}} \sum_{u=1}^H \sum_{v=1}^W \mathbf{M}_{\text{qry}, i}(u, v) + \epsilon}$
            \State Accumulate task meta-gradient: $\nabla_\Theta \mathcal{L} \leftarrow \nabla_\Theta \mathcal{L} + \frac{1}{B} \nabla_\Theta \mathcal{L}_{\text{qry}}$
        \EndFor
        
        \State Clip gradients: $\|\nabla_\theta\|_2 \le 1.0$, \quad $\|\nabla_\phi\|_2 \le 1.0$
        \State Update parameters: $\Theta \leftarrow \text{AdamW}(\Theta, \nabla_\Theta \mathcal{L}, \eta, \text{weight\_decay}=10^{-5})$
    \EndFor
    
    \State \Return Optimized parameters $\theta^*, \phi^*$
\end{algorithmic}
\end{algorithm}

\section{Testing Algorithm}
\label{sec:test_algo}
\begin{algorithm}[H]
\caption{\textsc{MetaLearnNCA} Few-Shot Offline Inference and Testing Loop}
\label{alg:inference}
\begin{algorithmic}[1]
\Require Optimized Active-NCA parameters $\theta^*$, Meta-NCA parameters $\phi^*$
\Require Target test dataset $\mathcal{D}_{\text{test}}$, Number of classes $N_{\text{way}}$, Support shot $K_{\text{shot}}$
\Require Base recurrence steps $N = 19$, Optimization repeats $K_{\text{repeats}} = 5$
\Require Optional test-time program spatial displacement $\Delta_{\text{shift}} \ge 0$ (default: 0)

\State Sample test episode: $(X_{\text{supp}}, y_{\text{supp}}), (X_{\text{qry}}, y_{\text{qry}}) \sim \mathcal{D}_{\text{test}}$ \Comment{$B_{\text{supp}} = N_{\text{way}} \times K_{\text{shot}}$}
\State Compute foreground stroke masks: 
\State \quad $M_{\text{supp}} \leftarrow \mathbb{I}(X_{\text{supp}} > -0.8)$, \quad $M_{\text{qry}} \leftarrow \mathbb{I}(X_{\text{qry}} > -0.8)$
\State Initialize random spatial program grid: $S_{\text{task}}^{(0)} \sim \mathcal{U}(0, 1)^{1 \times C_s \times H \times W}$

\Statex \vspace{-1mm} 
\State $S_{\text{eval}}^{(0)} \leftarrow \operatorname{roll}\left(S_{\text{task}}^{(0)}, (\Delta_{\text{shift}}, \Delta_{\text{shift}})\right)$
\State $S_{\text{qry}} \leftarrow \operatorname{expand}\left(S_{\text{eval}}^{(0)}, B_{\text{qry}}\right)$
\State $D_{\text{qry}} \leftarrow f_{\theta^*}\left(X_{\text{qry}}, S_{\text{qry}}, \text{steps} = 2N\right)$
\State $\hat{y}_{\text{qry}, i} \leftarrow \arg\max_{c \in \{0, \dots, C_{\text{loss}}-1\}} \sum_{u=1}^H \sum_{v=1}^W D_{\text{qry}, i}^{[c, u, v]} \cdot M_{\text{qry}, i}^{[u, v]} \quad \forall i \in \{1, \dots, B_{\text{qry}}\}$
\State $\text{Acc}^{(0)} \leftarrow \frac{1}{B_{\text{qry}}} \sum_{i=1}^{B_{\text{qry}}} \mathbb{I}\left(\hat{y}_{\text{qry}, i} = y_{\text{qry}, i}\right) \times 100\%$
\Statex \vspace{-1mm} 
\For{$k = 1$ \textbf{to} $K_{\text{repeats}}$}
    \Statex \quad \textit{// 1. Forward Task Inference on Support Set (Active-NCA)}
    \State $S_{\text{supp}} \leftarrow \operatorname{expand}\left(S_{\text{task}}^{(k-1)}, B_{\text{supp}}\right)$
    \State $D_{\text{supp}} \leftarrow f_{\theta^*}\left(X_{\text{supp}}, S_{\text{supp}}, \text{steps} = 2N\right)$ \Comment{Forward unroll without gradients}
    
    \Statex \quad \textit{// 2. Signed Residual Error Calculation}
    \State Construct spatial one-hot targets $Y_{\text{spatial}} \in \{0, 1\}^{B_{\text{supp}} \times C_{\text{loss}} \times H \times W}$ from $y_{\text{supp}}$
    \State $E_{\text{pixel}} \leftarrow \left(Y_{\text{spatial}} - \operatorname{Softmax}\left(D_{\text{supp}}^{[0:C_{\text{loss}}]}\right)\right) \odot M_{\text{supp}}$
    
    \Statex \quad \textit{// 3. Cellular Optimization via Error Diffusion (Meta-NCA)}
    \State $D_{\text{meta}} \leftarrow g_{\phi^*}\left( E_{\text{pixel}}, D_{\text{meta}}, \text{steps} = N\right)$
    
    \Statex \quad \textit{// 4. Cross-Class Spatial Program Superposition}
    \State $S_{\text{task}}^{(k)} \leftarrow \frac{1}{B_{\text{supp}}} \sum_{i=1}^{B_{\text{supp}}} D_{\text{meta}, i}^{[0:C_s, :, :]}$ 
    
    \Statex \quad \textit{// 5. Query Set Evaluation (Gradient-Free Inference)}
    \State $S_{\text{eval}}^{(k)} \leftarrow \operatorname{roll}\left(S_{\text{task}}^{(k)}, (\Delta_{\text{shift}}, \Delta_{\text{shift}})\right)$ \Comment{Optional coordinate shift test}
    \State $S_{\text{qry}} \leftarrow \operatorname{expand}\left(S_{\text{eval}}^{(k)}, B_{\text{qry}}\right)$
    \State $D_{\text{qry}} \leftarrow f_{\theta^*}\left(X_{\text{qry}}, S_{\text{qry}}, \text{steps} = 2N\right)$
    \State $\hat{y}_{\text{qry}, i} \leftarrow \arg\max_{c} \sum_{u, v} \left(D_{\text{qry}, i}^{[c, u, v]} \cdot M_{\text{qry}, i}^{[u, v]}\right) \quad \forall i \in \{1, \dots, B_{\text{qry}}\}$
    \State $\text{Acc}^{(k)} \leftarrow \frac{1}{B_{\text{qry}}} \sum_{i=1}^{B_{\text{qry}}} \mathbb{I}\left(\hat{y}_{\text{qry}, i} = y_{\text{qry}, i}\right) \times 100\%$
\EndFor

\State \textbf{return} Adaptation trajectory $[\text{Acc}^{(0)}, \text{Acc}^{(1)}, \dots, \text{Acc}^{(K_{\text{repeats}})}]$ and adapted program $S_{\text{task}}^{(K_{\text{repeats}})}$
\end{algorithmic}
\end{algorithm}

\section{Complete Channel Atlases}
\label{app:atlases}

\begin{figure}[H]
    \centering
    \includegraphics[width=0.85\linewidth]{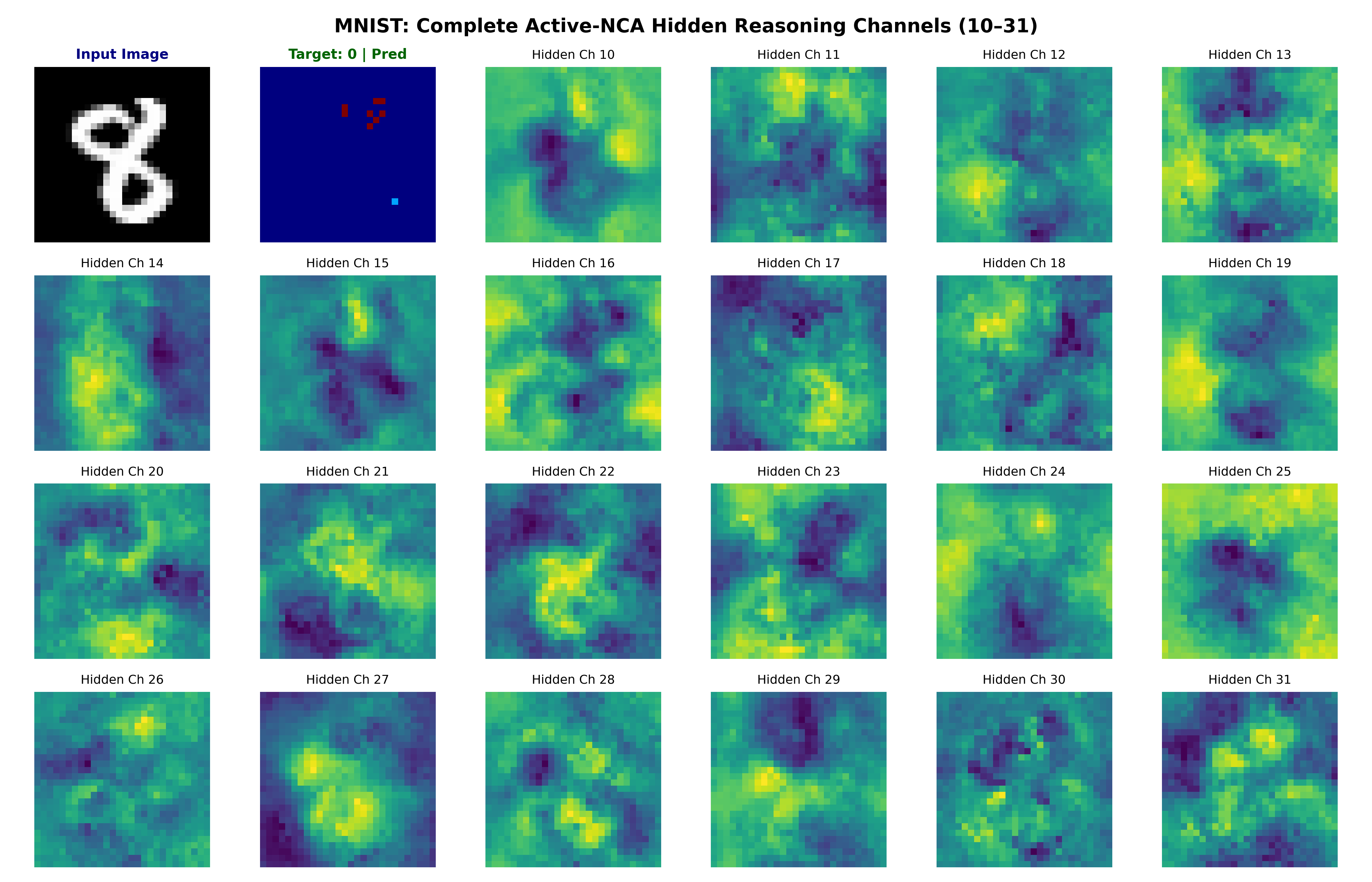}
    \caption{Complete atlas of Active-NCA dynamic reasoning channels (Channels 10 through 31) during MNIST inference.}
    \label{fig:hidden_vis_full}
\end{figure}

\begin{figure}[H]
    \centering
    \includegraphics[width=0.85\linewidth]{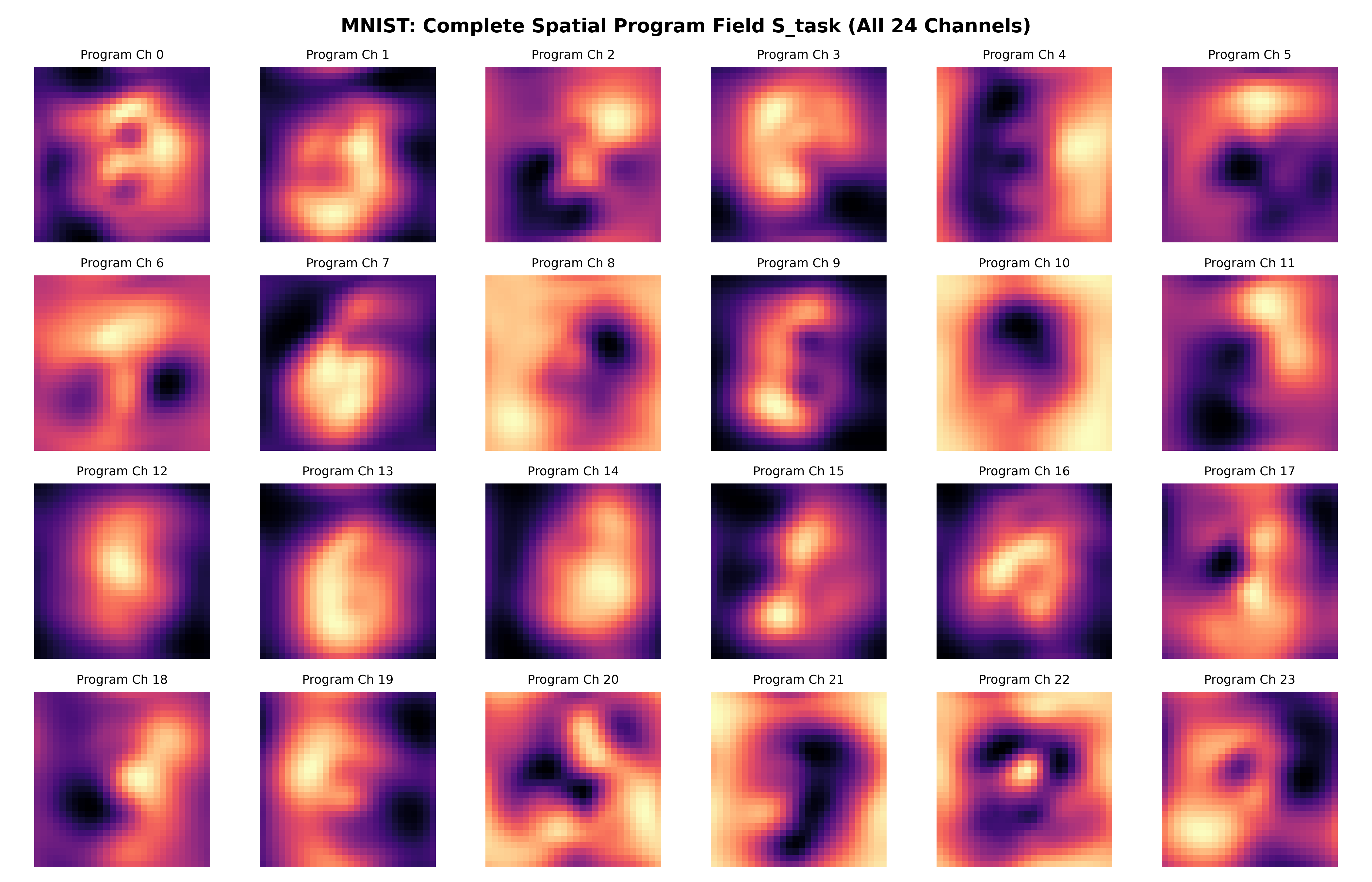}
    \caption{Complete atlas of static spatial program channels ($S_{\text{task}}$, Channels 0 through 23) written by the Meta-NCA following support adaptation.}
    \label{fig:prog_viz_full}
\end{figure}

\section{Hidden Channel and Spatial Program Ablation}
\label{app:hcspa}
To understand how task representations and decision-making are distributed across the cellular state channels, we conduct targeted channel ablation experiments on the Omniglot test set.

\begin{figure}[t]
    \centering
    \begin{subfigure}[b]{0.48\linewidth}
        \centering
        \includegraphics[width=\linewidth]{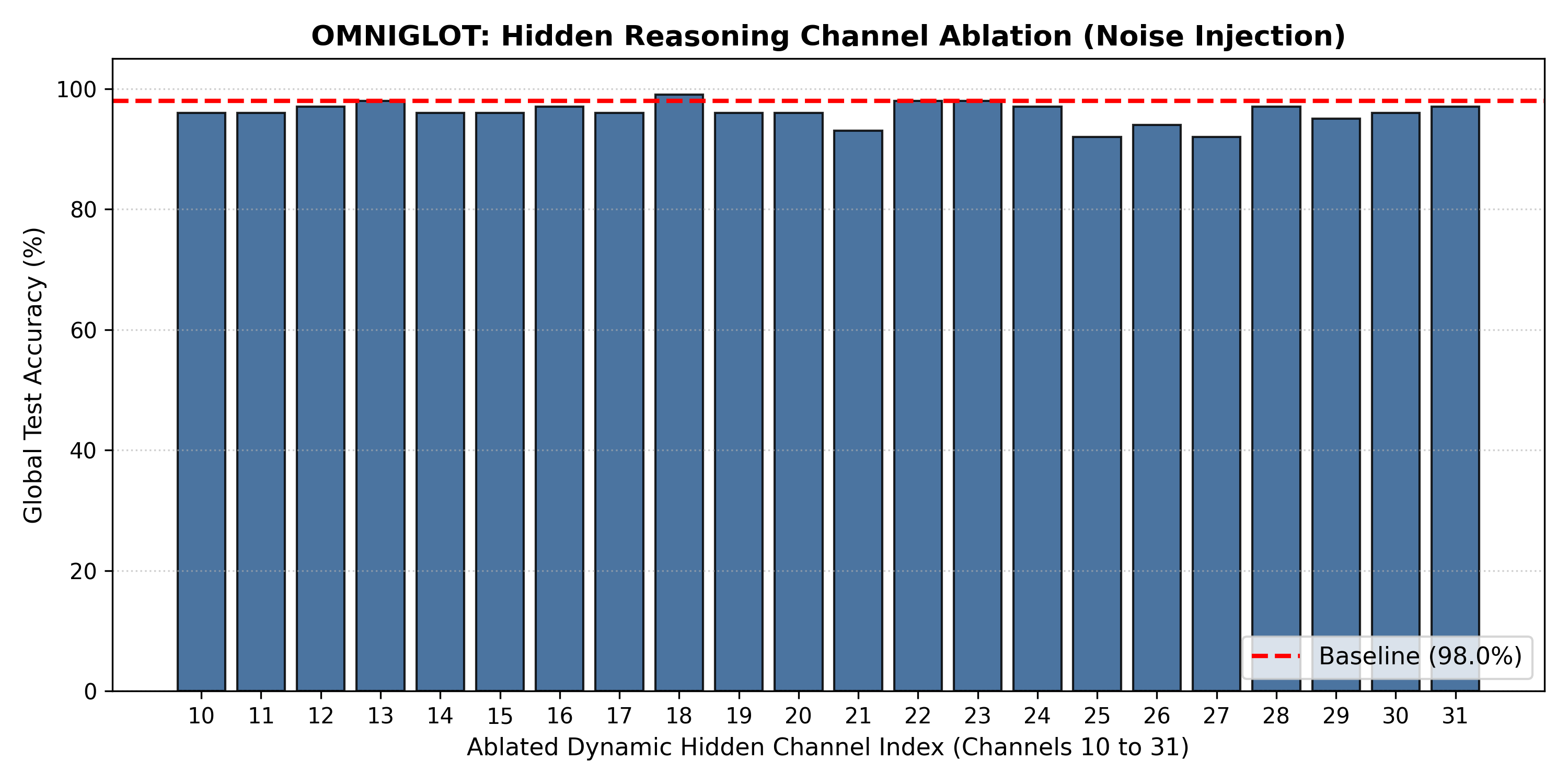}
        \caption{Active-NCA Dynamic Reasoning Channels}
        \label{fig:hidden_abb}
    \end{subfigure}
    \hfill
    \begin{subfigure}[b]{0.48\linewidth}
        \centering
        \includegraphics[width=\linewidth]{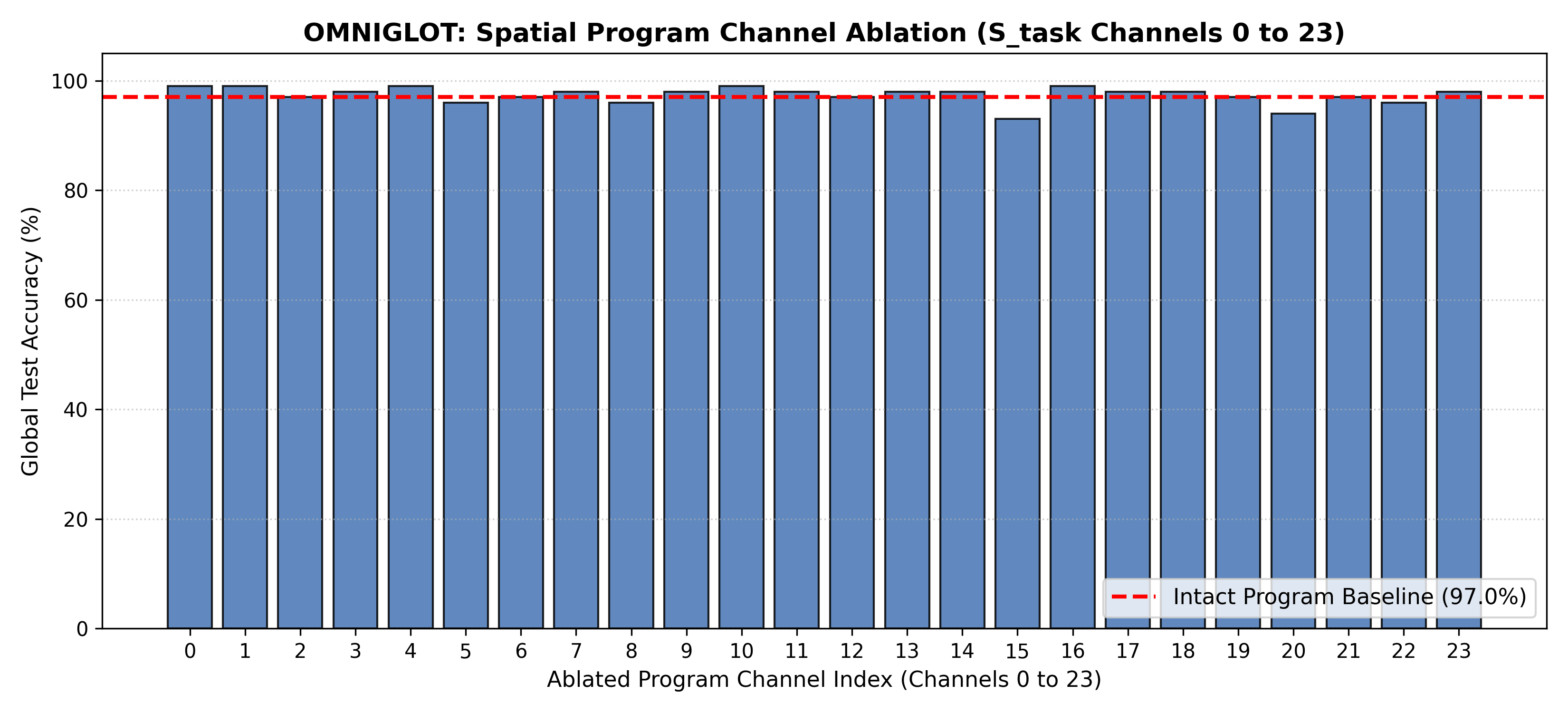}
        \caption{Meta-NCA Static Program Channels}
        \label{fig:prog_abb}
    \end{subfigure}
    \caption{Ablation analysis on Omniglot: (a) test accuracy under Gaussian noise injection across dynamic reasoning channels ($10 \dots 31$), and (b) test accuracy when individual static program channels ($0 \dots 23$) are ablated to zero.}
    \label{fig:ablation_joint}
\end{figure}

\paragraph{Active-NCA Dynamic Reasoning Channels (Figure~\ref{fig:hidden_abb}):} 
We inject independent Gaussian noise ($\mathcal{N}(0, 1)$) into individual dynamic reasoning channels ($c \in \{10, \dots, 31\}$) throughout test-time execution while keeping all other channels intact. As shown in Figure~\ref{fig:hidden_abb}, the unablated baseline achieves $\mathbf{98.0\%}$ accuracy. Ablating individual reasoning channels results in graceful degradation: the majority of channels maintain accuracy above $94\%$, with only specific channels (e.g., channels 20, 22, and 26) causing moderate drops to $\approx 92\%$--$94\%$. Accuracy is relatively insensitive to individual channel ablation. The higher baseline accuracy compared to the reported main results is due to testing variance, as these results are only tested on one seed. 

\paragraph{Meta-NCA Spatial Program Channels (Figure~\ref{fig:prog_abb}):} 
Similarly, we ablate each of the $C_s = 24$ static spatial program channels ($S_{\text{task}}$) by setting them to zero following support adaptation. Figure~\ref{fig:prog_abb} reveals that the baseline ($97.0\%$) drops by at most $1.5\%$--$3.0\%$ when any single program channel is ablated. This demonstrates that the Meta-NCA generates redundant, multi-channel spatial pre-activation codes.

\section{Flexible Program}
\label{app:fp}

\begin{figure}[H]
    \centering
    \includegraphics[width=1.\linewidth]{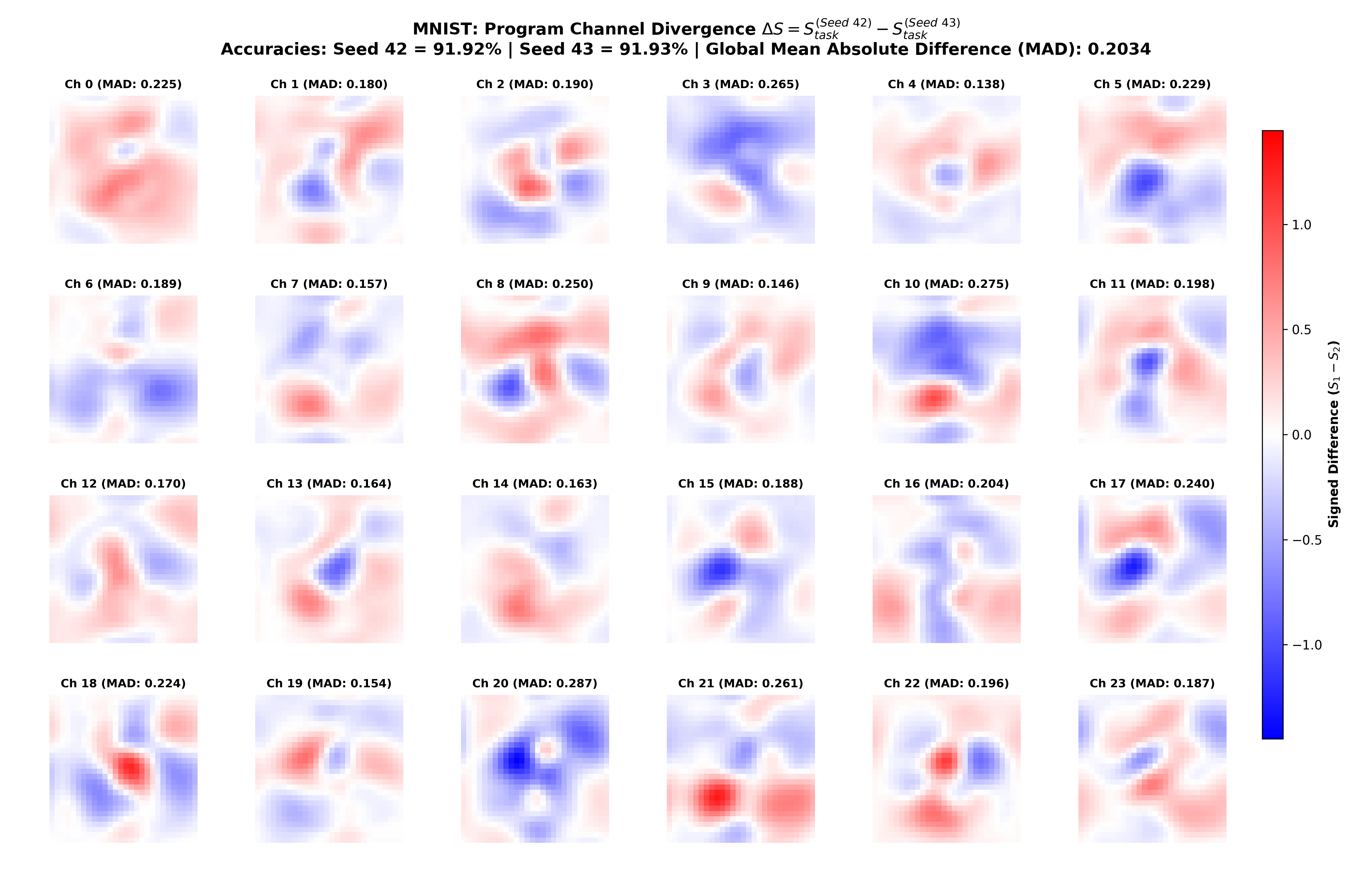}
    \caption{Difference in $S_{task}$ across 2 independent testing runs on the same training seed}
    \label{fig:program_diff}
\end{figure}

\paragraph{Solution Manifold and Functional Degeneracy.}
To inspect the stability and uniqueness of the learned spatial programs, we evaluate a single trained model checkpoint (\texttt{Seed 0}) across two independent test-time evaluation runs (\texttt{Seed 42} vs.\ \texttt{Seed 43}) on the MNIST benchmark. In this setting, the model parameters $\{\theta, \phi\}$ remain identical, but the evaluation runs differ across three stochastic dimensions: (i) each run samples an independent 10-shot support set with differing handwriting styles, (ii) the static program grid is initialized from distinct random noise $S_{\text{task}}^{(0)} \sim \mathcal{U}(0,1)$, and (iii) cellular unrolling is driven by independent stochastic update masks $M_{\text{stoch}} \sim \text{Bernoulli}(0.5)$.

As visualized in Figure~\ref{fig:program_diff}, both runs achieve near identical downstream test accuracy across all $10{,}000$ MNIST test samples ($91.92\%$ vs.\ $91.93\%$). Subtracting the two adapted programs channel-by-channel ($\Delta S = S_{\text{task}}^{(\text{Seed 42})} - S_{\text{task}}^{(\text{Seed 43})}$) reveals a global Mean Absolute Difference of $\text{MAD} = 0.2034$, with localized regions differing by up to $\pm 1.2$. This divergence is not random high-frequency static; rather, it manifests as smooth, spatially continuous dipoles and phase shifts. This behavior highlights two core dynamical properties:

 \textbf{Support-Exemplar Adaptation without Overfitting:} The localized variations in $S_{\text{task}}$ reflect local adjustments to the specific stylistic traits (stroke thickness, slant) of the respective support sets, while remaining constrained enough to preserve macro-level class separability across the entire test set.
 
\textbf{Functional Degeneracy:} The presence of non-trivial spatial divergence alongside identical functional performance demonstrates that the Meta-NCA optimization landscape does not collapse to an isolated, fragile minimum. This Mirrors biological degeneracy, where distinct cellular micro-states reliably execute equivalent computational phenotypes.

\section{Simulated spatial-memory corruption}
\label{app: ssmc}

\begin{figure}[H]
    \centering
    \includegraphics[width=0.85\linewidth]{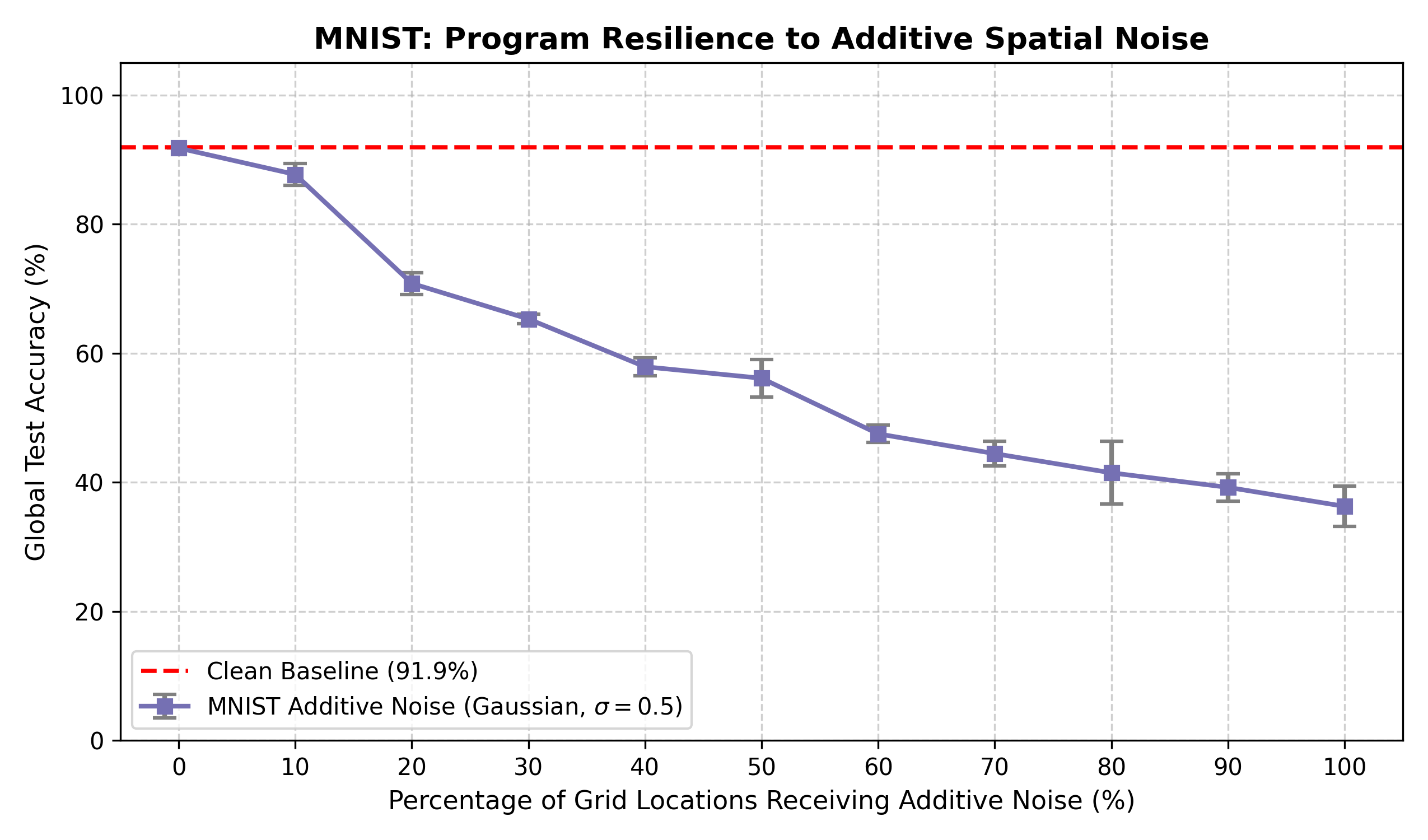}
    \caption{MetaLearnNCA performance on MNIST under different levels of additive random noise to $S_{task}$.}
    \label{fig:addrandonoise}
\end{figure}
\begin{figure}[H]
    \centering
    \includegraphics[width=0.85\linewidth]{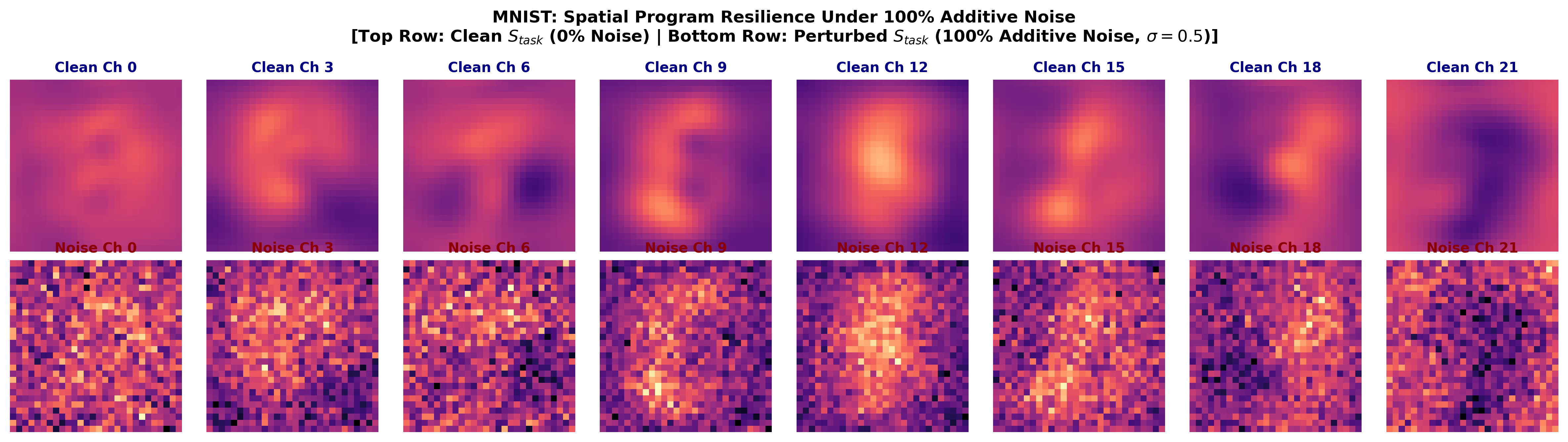}
    \caption{MetaLearnNCA $S_{task}$ under different levels of noise.}
    \label{fig:addrandonoise_show}
\end{figure}

\paragraph{Visualizing Additive Noise Robustness.}
To provide intuitive grounding for the noise resilience curve in Figure~\ref{fig:addrandonoise}, Figure~\ref{fig:addrandonoise_show} compares the clean spatial program grid against a grid corrupted by additive Gaussian noise ($\sigma = 0.5$) across $100\%$ of spatial coordinates. As visible in the bottom row, this perturbation severely degrades local pixel fidelity, simulating high-amplitude thermal or transmission noise in physical analog substrates. Crucially, the global low-frequency modes of the program remain somewhat structurally coherent. 

\section{Dataset to Dataset Transfer}
\label{app:d2dt}
\begin{figure}[H]
    \centering
    \includegraphics[width=0.85\linewidth]{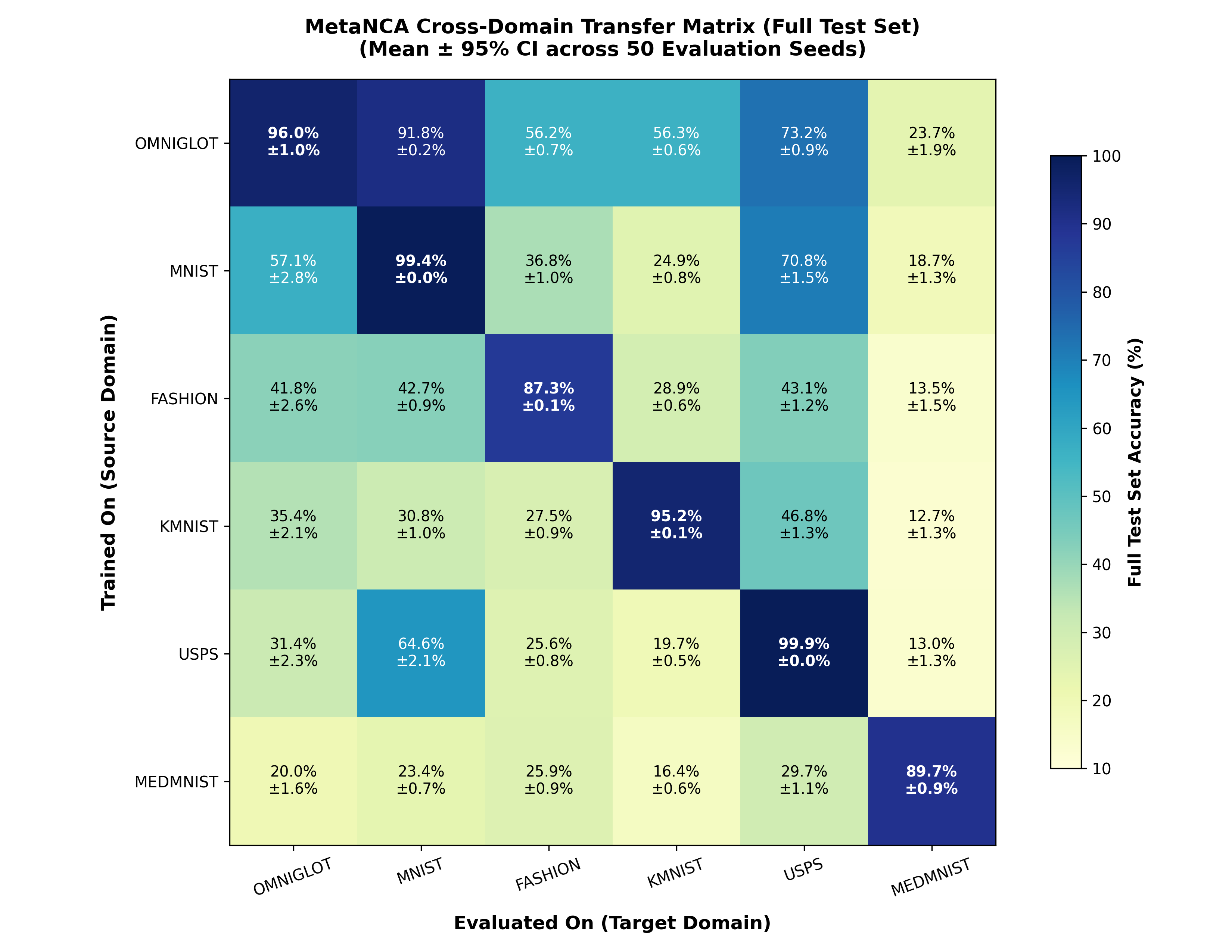}
    \caption{Cross dataset transfer, each model is trained on one dataset and tested on all the other ones. 50 repeats per dataset with 95\% CI intervals.}
    \label{fig:d2dtransfer}
\end{figure}

\section{K-repeats and K-shot meta-learning surface}
\label{app:kakmls}

\begin{figure}[H]
    \centering
    \includegraphics[width=0.85\linewidth]{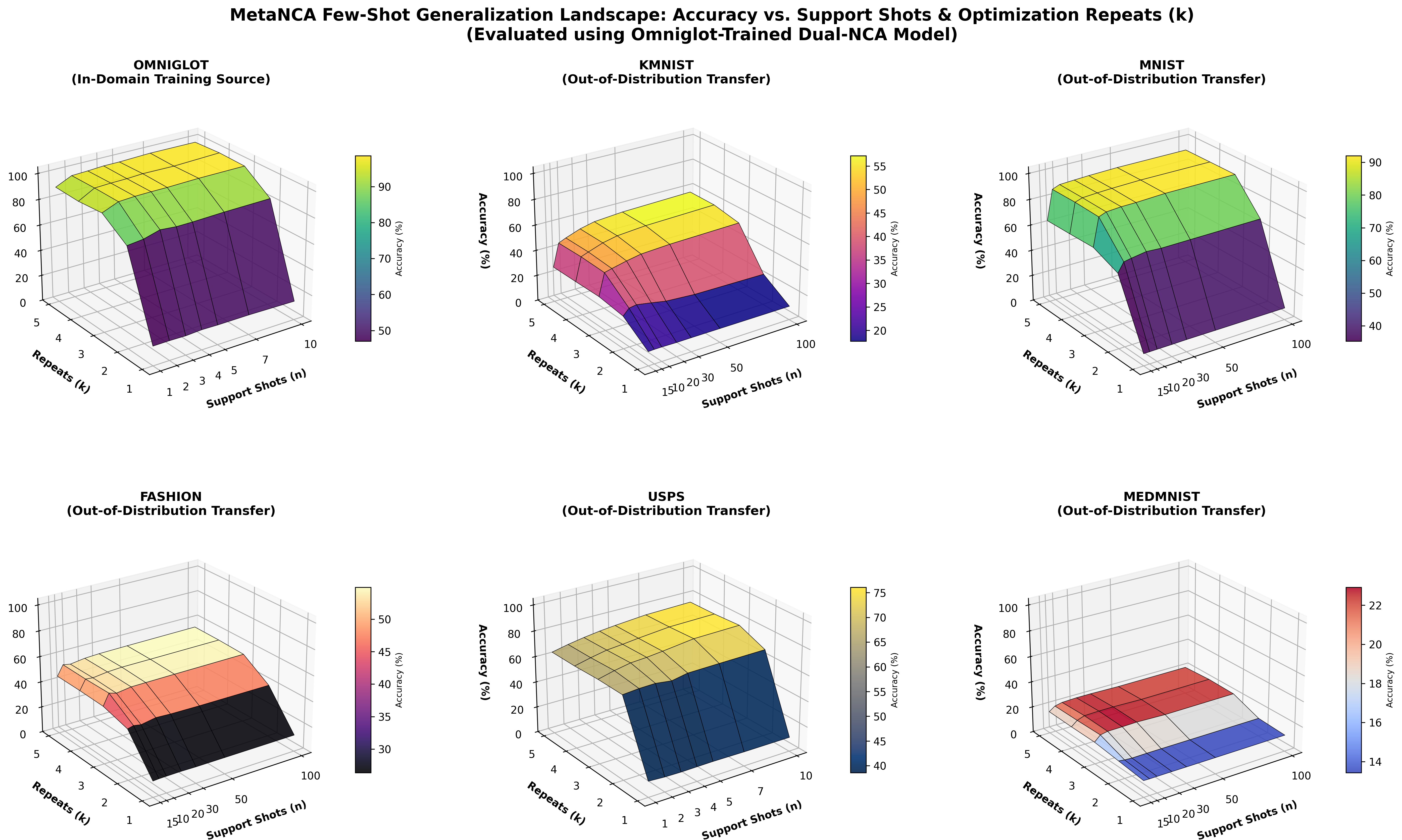}
    \caption{Dataset surfaces for seed 0, trained on OMNIGLOT across a combination of K-repeats and K-shots.}
    \label{fig:d2dtransfer_surf}
\end{figure}

\section{Spatial Translation and Coordinate Invariance Robustness}
\label{sec:spatial_translation_robustness}

A core design characteristic of \textsc{MetaLearnNCA} is that task knowledge is encoded as a continuous 2D spatial field ($S_{\text{task}} \in \mathbb{R}^{C_s \times H \times W}$) rather than a global 1D latent vector. While this inductive bias provides localized routing mechanisms for stroke- and shape-based reasoning, it introduces a natural question: \emph{does the coupled cellular dynamics rely on rigid pixel-to-pixel coordinate alignment, or does it exhibit translational elasticity under spatial displacement?}

\subsection{Experimental Setup}
To quantify spatial coordinate robustness, we evaluate a model meta-trained on Omniglot (Seed 0) in a standard 5-shot, 10-way evaluation setting across all six benchmark domains. Following support-set adaptation on canonical images, we apply independent discrete 2D spatial translations to all query instances:
\begin{equation}
X_{\text{shifted}}(u, v) = X(u + \Delta_u, v + \Delta_v), \quad 
\end{equation}
with displacement thresholds $\Delta \in \{1, 2, 3, 4\}$ pixels, using circular padding to preserve canvas dimensions ($H = W = 28$). We evaluate classification accuracy across 5 independent evaluation runs per displacement setting, and report sample means with $95\%$ confidence intervals.

\begin{table}[h]
\centering
\caption{Robustness of \textsc{MetaLearnNCA} to discrete spatial coordinate shifts ($\Delta \in \{1, 2, 3, 4\}$ pixels) in the 5-shot, 10-way transfer regime. Results report mean classification accuracy $\pm$ 95\% confidence intervals evaluated over 5 independent testing runs on full test splits.}
\label{tab:spatial_translation_robustness}
\vspace{2mm}
\small
\setlength{\tabcolsep}{7pt}
\begin{tabular}{l cccc}
\toprule
\textbf{Dataset} & \textbf{Shift = 1 px} & \textbf{Shift = 2 px} & \textbf{Shift = 3 px} & \textbf{Shift = 4 px} \\
\midrule
\textbf{Omniglot (In-Dist)} & $93.47 \pm 4.51\%$ & $90.93 \pm 4.56\%$ & $83.33 \pm 4.86\%$ & $70.13 \pm 8.81\%$ \\
\textbf{MNIST}              & $84.43 \pm 1.41\%$ & $82.93 \pm 2.62\%$ & $77.13 \pm 2.86\%$ & $59.01 \pm 3.37\%$ \\
\textbf{USPS}               & $57.89 \pm 7.27\%$ & $51.67 \pm 6.85\%$ & $40.71 \pm 5.31\%$ & $30.83 \pm 2.51\%$ \\
\textbf{Fashion-MNIST}      & $48.72 \pm 4.56\%$ & $42.85 \pm 4.01\%$ & $34.03 \pm 5.16\%$ & $24.72 \pm 4.66\%$ \\
\textbf{KMNIST}             & $41.70 \pm 3.88\%$ & $37.47 \pm 4.02\%$ & $30.50 \pm 3.55\%$ & $22.73 \pm 3.75\%$ \\
\textbf{MedMNIST}           & $18.55 \pm 8.05\%$ & $17.23 \pm 7.92\%$ & $14.98 \pm 6.98\%$ & $13.00 \pm 5.90\%$ \\
\bottomrule
\end{tabular}
\end{table}

\subsection{Results and Dynamical Elasticity}
As summarized in Table~\ref{tab:spatial_translation_robustness}, \textsc{MetaLearnNCA} displays graceful, non-catastrophic degradation across all evaluated visual domains:

\begin{enumerate}
    \item \textbf{Near-Invariance at Local Offsets ($\Delta \le 2$ px):} For local shifts of $\Delta = 1$ pixel, accuracy remains virtually indistinguishable from clean unshifted baselines (e.g., MNIST achieves $84.43\% \pm 1.41\%$, and Omniglot attains $93.47\% \pm 4.51\%$). At $\Delta = 2$ pixels, degradation remains minimal, with MNIST dropping by only $1.50\%$ ($82.93\%$) when compared to $\Delta = 1$ and in-distribution Omniglot retaining $>90.9\%$ mean accuracy.
    \item \textbf{Degradation under Larger Offsets ($\Delta = 4$ px):} When images are perturbed by $\Delta = 4$ pixels, representing a larger displacement exceeding $14.2\%$ of the total canvas width ($28 \times 28$), the model retains some task structure. On MNIST, the model achieves $59.01\% \pm 3.37\%$, which is nearly $6\times$ higher than the $10.0\%$ random-guess baseline. Similarly, Omniglot retains $70.13\% \pm 8.81\%$.
\end{enumerate}

\paragraph{Mechanistic Basis of Elasticity.} This shows that $S_{\text{task}}$ does not solely function as an overfitted, static pixel mask. Rather, some degree of translational elasticity is an emergent consequence of the Active-NCA's recurrent message-passing dynamics. Because each update step convolves differential operators ($K_{\text{sobel}}, K_{\text{lap}}$) across a $3 \times 3$ neighborhood, information propagates outwards at 1 pixel per step. Across forward inference steps, each cell integrates context over an effective receptive field spanning the entire canvas. This recurrent diffusion acts as an intrinsic spatial regularizer: local gradients dynamically warp and re-route displaced stroke trajectories into the attractor basin defined by $S_{\text{task}}$, buffering the system against some coordinate misalignment.

\section{Baseline Hyperparameters and Dataset Preprocessing}
\label{sec:baseline_hyperparameters}

All canonical baselines share a standard \texttt{Conv4-64} encoder (4 conv blocks with 64 channels, ReLU, $2\times 2$ max-pooling, and global average pooling yielding a 64D feature vector). All models are meta-trained for 5,000 epochs on Omniglot with AdamW ($\text{lr} = 10^{-3}$, weight decay $= 10^{-5}$, gradient clipping at 1.0). Full source code and configuration files are available in our anonymous repository.\footnote{\url{https://github.com/etimush/MetaLearnNCA}}

\begin{table}[h]
\centering
\caption{Architectural and optimization hyperparameters for canonical baselines.}
\label{tab:baseline_hyperparams_compact}
\vspace{1.5mm}
\resizebox{\textwidth}{!}{%
\begin{tabular}{l cccc}
\toprule
\textbf{Hyperparameter} & \textbf{ProtoNet} & \textbf{MatchingNet} & \textbf{RelationNet} & \textbf{FOMAML} \\
\midrule
Backbone & Conv4-64 & Conv4-64 & Conv4-64 & Conv4-64 \\
Parameters & $\approx 115\text{k}$ & $\approx 115\text{k}$ & $\approx 225\text{k}$ & $\approx 115\text{k}$ \\
Output Head & Metric Centroid & Cosine Attention & Relation MLP & Linear ($64 \to 10$) \\
BatchNorm Stats & Frozen Running & Frozen Running & Frozen Running & Shuffled Mini-Batch \\
Meta-Loss & Cross-Entropy & Neg. Log-Likelihood & Mean Squared Error & Cross-Entropy \\
Inner Steps ($K_{\text{train}} / K_{\text{test}}$) & -- & -- & -- & 3 / 5 \\
Inner LR ($\alpha_{\text{train}}$) & -- & -- & -- & 0.1 (clip = 1.0) \\
In-Domain Inner LR ($\alpha_{\text{omni}}$) & -- & -- & -- & 0.2 ($\le 5$-shot), 0.1 (10-shot) \\
OOD Inner LR ($\alpha_{\text{ood}}$) & -- & -- & -- & 0.1 (1s), 0.05 (5s), 0.03 (10s) \\
Higher-Order Gradients & -- & -- & -- & First-Order (FOMAML) \\
\bottomrule
\end{tabular}%
}
\end{table}

\begin{table}[h]
\centering
\caption{Dataset preprocessing and normalization pipelines ($28 \times 28$ grayscale).}
\label{tab:transforms_compact}
\vspace{1.5mm}
\resizebox{\textwidth}{!}{%
\begin{tabular}{lll}
\toprule
\textbf{Dataset} & \textbf{Domain} & \textbf{Transform Pipeline} \\
\midrule
\textbf{Omniglot} & Sparse Characters (In-Dist) & $\operatorname{Resize}(28) \to \operatorname{ToTensor}() \to \text{Invert } (1.0-x) \to \operatorname{Normalize}(0.5, 0.5)$ \\
\textbf{MNIST / KMNIST / USPS} & Digits / Calligraphy (OOD) & $\operatorname{Resize}(28) \to \operatorname{ToTensor}() \to \operatorname{Normalize}(0.5, 0.5)$ \\
\textbf{Fashion / MedMNIST} & Silhouettes / CT Organs (OOD) & $\operatorname{Resize}(28) () \to \operatorname{ToTensor}() \to \operatorname{Normalize}(0.5, 0.5)$ \\
\bottomrule
\end{tabular}%
}
\end{table}

\end{document}